%% file: main.tex
\documentclass[11pt]{article}
\PassOptionsToPackage{table,dvipsnames}{xcolor}
\PassOptionsToPackage{numbers,sort&compress}{natbib}
\usepackage{amsmath,amssymb}
\usepackage{booktabs,longtable}
\usepackage{hyperref}
\usepackage{graphicx}
\usepackage{tikz}
\usetikzlibrary{arrows.meta,positioning,fit,shapes.multipart,backgrounds,decorations.pathreplacing}
\usepackage{enumitem}
\usepackage{array}
\usepackage{float}
\usepackage{caption}
\usepackage[textsize=small]{todonotes}
\usepackage{xcolor}
\usepackage{bm}

\usepackage[utf8]{inputenc} %
\usepackage[T1]{fontenc}    %
\usepackage{url}            %
\usepackage{amsfonts}       %
\usepackage{nicefrac}       %
\usepackage{microtype}      %
\usepackage{times}
\usepackage{style/perception_conference}
\iclrfinalcopy
\usepackage{epsfig}
\usepackage{amsmath}
\usepackage{makecell} 
\usepackage{tabularx}
\usepackage{diagbox}
\usepackage{threeparttable}

\usepackage{xspace}
\usepackage{subcaption}
\usepackage{multirow} 
\usepackage{colortbl}
\usepackage{pifont}
\usepackage{wrapfig}
\usepackage{tcolorbox}
\tcbuselibrary{skins,breakable}
\newtcolorbox{examplecard}[1][]{enhanced, breakable, colback=white, colframe=black!10,
  boxrule=0.35pt, arc=2pt, left=4pt, right=4pt, top=1pt, bottom=2pt,
  #1}
\usepackage{marvosym}
\usepackage{adjustbox}  

\newtcbox{\hlpill}[1][]{nobeforeafter, tcbox raise base, enhanced, coltext=black,
  colback=black!3, colframe=black!15, boxsep=0.7pt, left=3pt, right=3pt,
  top=0.4pt, bottom=0.4pt, arc=2pt, boxrule=0.4pt, #1}

\newtcbox{\hlbox}[1][]{nobeforeafter, tcbox raise base, enhanced, coltext=black,
  colback=Emerald!10, colframe=Emerald!40!black, boxsep=0.7pt, left=3pt, right=3pt,
  top=0.4pt, bottom=0.4pt, arc=0pt, boxrule=0.5pt, #1}

\newcommand{\hldelta}[1]{\fboxsep=1.5pt\colorbox{Goldenrod!30}{$#1$}}
\newcommand{\hldeltabox}[1]{\fboxsep=1.5pt\colorbox{Goldenrod!30}{#1}}
\newcommand{\hlmath}[1]{\tikz[baseline=(X.base)]{\node[fill=Goldenrod!30,inner sep=1.5pt](X){$\displaystyle #1$};}}

\usepackage{listings}
\usepackage{upquote} 
\usepackage{algorithm}
\usepackage[noend]{algpseudocode}
\makeatletter
\algnewcommand{\LineComment}[1]{\Statex \hskip\ALG@thistlm $\triangleright$\, #1}
\algrenewcommand{\algorithmiccomment}[1]{\hfill\textcolor{gray}{\footnotesize$\triangleright$\, #1}}
\makeatother

\setcitestyle{numbers,square,sort&compress}

\newcommand{\perceptronlogo}[1][0.26\linewidth]{\includegraphics[width=#1]{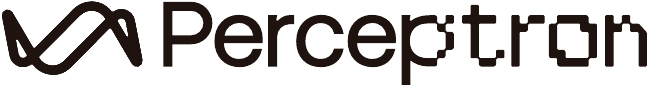}}

\input{figures}

\title{Improving MoE Compute Efficiency by Composing Weight and Data Sparsity}
\author{
  Maciej Kilian \And Oleg Mkrtchyan \And Luke Zettlemoyer{\NoHyper\thanks{University of Washington}\endNoHyper} \AND Akshat Shrivastava \And Armen Aghajanyan
}

\titlegraphic{\perceptronlogo[0.30\linewidth]}

\begin{document}
\raggedbottom
\maketitle

\begin{abstract}
Mixture-of-Experts layers achieve compute efficiency through weight sparsity: each token activates only a subset of experts. Data sparsity, where each expert processes only a subset of tokens, offers a complementary axis. Expert-choice routing implements data sparsity directly but violates causality in autoregressive models, creating train-inference mismatch. We recover data sparsity within causal token-choice MoE by leveraging zero-compute (null) experts within the routing pool. When a token routes to null experts, those slots consume no compute. The standard load balancing objective trains the model to uniformly use all experts (real and null) therefore creating data sparsity in expectation without the causality violations. We evaluate on vision-language model training, where data heterogeneity is pronounced: vision encoders produce many low-information tokens while text tokens are denser. At matched expected FLOPs, composing weight and data sparsity yields a more compute-efficient frontier than weight sparsity alone, with gains in training loss and downstream performance. The model learns implicit modality-aware allocation, routing vision tokens to null experts more aggressively than text, without explicit modality routing.

\vspace{-0.6em}\end{abstract}

\section{Introduction}
Mixture-of-Experts (MoE) layers \cite{fedus2022switchtransformersscalingtrillion, lepikhin2020gshardscalinggiantmodels} have enabled more efficient scaling of Transformers through weight sparsity: they replicate the FFN into many experts and use a router to select a small subset per token, enabling conditional computation with large parameter counts while keeping per-token compute fixed. But data itself is heterogeneous and redundant. Many inputs contain large low-information regions-blank or repetitive patches, punctuation, boilerplate, predictable continuations-suggesting that not every token deserves the same compute budget.

This motivates data sparsity: instead of allocating weights per token, allocate tokens per weight. Weight and data sparsity are dual views of the same routing matrix $R \in \{0,1\}^{T \times N}$: weight sparsity constrains columns (each token activates at most $k$ experts), while data sparsity constrains rows (each expert processes a bounded number of tokens). Composing both yields a budget over the full token$\times$expert matrix.

The challenge is implementing data sparsity in autoregressive models. Expert-choice MoE \cite{zhou2022mixtureofexpertsexpertchoicerouting}  is data-sparse by design-each expert selects its tokens-which achieves weight sparsity in expectation since expert overlap is unlikely. However, this requires access to future tokens, creating non-causal dependencies and train-inference mismatch. Token-choice MoE \cite{fedus2022switchtransformersscalingtrillion} preserves causality by letting each token select its experts independently, but assigns exactly $k$ experts to every token regardless of information content.

We recover data sparsity within causal token-choice MoE by leveraging zero-compute (null) experts~\cite{jin2024moeacceleratingmixtureofexpertsmethods, zeng2024adamoetokenadaptiveroutingnull, meituanlongcatteam2025longcatflashtechnicalreport} to the routing pool. When the router assigns a token to null experts, those slots skip expert computation. The standard load balancing objective \cite{qiu2025demonsdetailimplementingload, fedus2022switchtransformersscalingtrillion} trains the model to uniformly use all experts-real and null-creating data sparsity in expectation without causality violations.

Data heterogeneity is particularly pronounced in multimodal training: vision encoders produce many tokens per image, most carrying little information, while text tokens tend to be denser. We therefore focus on vision-language model training as a setting where data sparsity should provide clear benefits. At matched expected FLOPs, configurations with data sparsity outperform those without, yielding gains in both training loss and downstream performance. The model learns implicit modality-aware allocation: vision tokens route to null experts more aggressively than text, shifting compute toward task-relevant information without explicit modality routing.

Our main contributions are:

\begin{itemize}[leftmargin=1.2em]

    \item Demonstration that composing weight and data sparsity improves the compute-efficiency frontier.

    \item A minimal modification to token-choice MoE that achieves this composition via zero-compute experts while preserving causality.

    \item Analysis showing modality-aware compute allocation emerges in multimodal training without explicit supervision.

\end{itemize}

\FigMethodOverview

\section{Related Work}
\paragraph{Mixture-of-Experts.} MoE architectures achieve compute efficiency through weight sparsity, activating a subset of experts per token. GShard~\citep{lepikhin2020gshardscalinggiantmodels} and Switch Transformer~\citep{fedus2022switchtransformersscalingtrillion} established token-choice top-K routing; expert-choice routing~\citep{zhou2022mixtureofexpertsexpertchoicerouting} inverts this by letting experts select tokens. Our work builds on token-choice routing but extends it to support data sparsity.

\paragraph{Multimodal MoE.} Vision-language models face modality imbalance: images produce many low-information tokens while text tokens are denser. V-MoE~\citep{riquelme2021scalingvisionsparsemixture} handles this via Batch Priority Routing; LiMoE~\citep{mustafa2022multimodalcontrastivelearninglimoe} uses an entropy-based regularization
scheme and shows modality-specific experts emerge organically; MoMa~\citep{lin2024momaefficientearlyfusionpretraining} partitions experts by modality architecturally. Our null expert mechanism achieves modality-aware allocation implicitly through standard load balancing.

\paragraph{Adaptive Computation.} Several approaches vary compute across tokens. Mixture-of-Depths~\citep{raposo2024mixtureofdepthsdynamicallyallocatingcompute} routes tokens to skip entire transformer blocks. MoE++~\citep{jin2024moeacceleratingmixtureofexpertsmethods} introduces zero-computation experts within MoE layers. AdaMoE~\citep{zeng2024adamoetokenadaptiveroutingnull} applies null experts in fine-tuning. LongCat-Flash~\citep{meituanlongcatteam2025longcatflashtechnicalreport} scales zero-compute experts to 560B parameters at 75\% data sparsity. Recent work on MoE router distribution shaping~\citep{mirvakhabova2025dirichletpriorshapingguidingexpert} incentivizes the router outputs to fit a specific probability distribution giving us more control over routing behavior.

We build on these approaches but emphasize a different lens: null experts compose weight sparsity (which experts) with data sparsity (which tokens), and this composition should only help-the solution space of denser configurations is preserved within sparser ones. Our contribution is compute-controlled experiments confirming this intuition: at matched expected FLOPs, weight-and-data-sparse MoE consistently outperforms weight-sparse-only baselines.

\section{Background}
\subsection{Weight and Data Sparsity}

Standard MoE \cite{fedus2022switchtransformersscalingtrillion, lepikhin2020gshardscalinggiantmodels} implements weight sparsity: each token activates $k$ of $N$ experts. Per-token compute is fixed at $k$ expert evaluations regardless of the token's information content.

Data sparsity varies compute across tokens. Some tokens use fewer than $k$ experts; others may use more. At aggregate level, expected compute remains controlled, but allocation adapts to the data.

As shown in Figure~\ref{fig:method_overview}, these dimensions are orthogonal. Consider the token$\times$expert routing matrix $R \in \{0,1\}^{T \times N}$ where $R_{t,e} = 1$ if token $t$ routes to expert $e$. Weight sparsity constrains columns: each token activates at most $k$ experts. Data sparsity constrains rows: each expert processes a bounded number of tokens. Composing both constrains the total budget $\sum_{t,e} R_{t,e}$.

This framing implies a natural experimental comparison. At matched expected compute, configurations with data sparsity ($\rho < 1$) can allocate 0 to $k$ experts per token, while dense configurations ($\rho = 1$) allocate exactly $k$ to every token. If composed sparsity outperforms at iso-compute, data sparsity provides value beyond what weight sparsity alone achieves.

\subsection{Token-Choice MoE}

A standard token-choice MoE \cite{fedus2022switchtransformersscalingtrillion} layer consists of $N$ expert networks $\mathcal{E} = \{E_1, \ldots, E_N\}$, a shared expert $E_{\text{shared}}$, and a router $G$ that activates the top-K experts per token:
\begin{equation}
    \bm{y} = E_{\text{shared}}(\bm{x}) + \sum_{i=1}^{N} g_i \cdot E_i(\bm{x}), \quad g_i =
    \begin{cases}
        \text{Softmax}(G(\bm{x}))_i & \text{if } i \in \text{top-K}(\{G(\bm{x})_j\}_{j=1}^N) \\
        0 & \text{otherwise}
    \end{cases}
\end{equation}
where $G(\bm{x}) = \bm{W}\bm{x}$ with $\bm{W} \in \mathbb{R}^{N \times D}$.

Token-choice routing preserves causality: each token selects its experts using only its own representation. This is essential for autoregressive models where future tokens are unavailable at inference.

\subsection{The Causality Challenge}

Expert-choice routing~\citep{zhou2022mixtureofexpertsexpertchoicerouting} implements data sparsity directly: each expert selects its top-K tokens from a batch, naturally allowing variable compute per token. However, this requires observing all tokens in a sequence simultaneously, violating causality in autoregressive models \cite{wang2024auxiliarylossfreeloadbalancingstrategy}. During training the model learns to rely on information from future tokens that will not be available at inference, creating train-inference mismatch.

Token-choice preserves causality but assigns exactly $k$ experts to every token, precluding data sparsity. We want both: the causality of token-choice and the variable allocation of expert-choice.
\section{Method}

We extend token-choice MoE with a minimal modification: adding null experts to the routing pool. This composes weight and data sparsity while preserving causality.

\subsection{Null Expert Mechanism}

We extend the router to output $N + 1$ logits by expanding $\bm{W} \in \mathbb{R}^{(N+1) \times D}$. The first $N$ logits correspond to real experts; the $(N+1)$th corresponds to the null expert, which outputs zero: $E_{\text{null}}(\bm{x}) = \bm{0}$. To control data sparsity, we duplicate the null logit $M$ times before top-K selection (\hldeltabox{highlighted} terms denote changes from token-choice):
\begin{equation}
    \tilde{G}(\bm{x}) = \left[ G(\bm{x})_{1:N},\ \underbrace{\hlmath{G(\bm{x})_{N+1}, \ldots, G(\bm{x})_{N+1}}}_{M \text{ copies}} \right] \in \mathbb{R}^{N+M}
\end{equation}
The gating function becomes:
\begin{equation}
    g_i =
    \begin{cases}
        \text{Softmax}(\hldelta{\tilde{G}}(\bm{x}))_i & \text{if } i \in \text{top-K}(\{\hldelta{\tilde{G}}(\bm{x})_j\}_{j=1}^{\hldelta{N+M}}) \text{ and } \hldelta{i \leq N} \\
        0 & \text{otherwise}
    \end{cases}
\end{equation}
Routing weights are renormalized over only the selected real experts, so output magnitude is unaffected by null routing.

\subsection{Notation}

We denote the number of real experts activated per token as the random variable $\mathbf{K}^{k_{\text{max}}}_{\rho}$, where $k_{\text{max}}$ is the maximum allocation (the top-K value) and $\rho \in (0,1]$ is the target data sparsity. The expected top-K is:
\begin{equation}
\label{eq:data_sparse_token_choice_notation}
    \mathbb{E}[\mathbf{K}^{k_{\text{max}}}_{\rho}] = k_{\text{max}} \cdot \rho
\end{equation}

For example, $\mathbf{K}^{8}_{0.5}$ denotes a configuration with top-8 routing at 50\% data sparsity, yielding $\mathbb{E}[\mathbf{K}^{8}_{0.5}] = 4$ expected real experts per token. For brevity, we refer to iso-compute families by their shared expectation, e.g., ``$\mathbb{E}[K] = 2$ runs.''

\paragraph{Model configurations.} We vary three dimensions: base model scale, maximum allocation $k_{\text{max}}$, and target data sparsity $\rho$. Configurations are denoted by base scale and routing parameters, e.g., 0.6B $\mathbf{K}^{8}_{0.5}$ indicates a 0.6B base model with $k_{\text{max}}=8$ at $\rho=0.5$. When we refer to parameter count alone (e.g., ``0.6B'' or ``1.7B''), we mean the base Qwen3 dense model~\cite{yang2025qwen3technicalreport} from which we initialize; when reporting full MoE scale we use the standard \texttt{[total]-A[active]} format, e.g., 5.3B-A1.2B denotes 5.3B total parameters with 1.2B active per token.

\subsection{Thresholding Interpretation}

Duplicating the null logit implements thresholding. For a token to activate exactly $r < k$ real experts, the router learns to set the null logit such that exactly $r$ real expert logits exceed it. The remaining $k - r$ slots go to null copies, contributing nothing. After renormalization, the output matches a standard top-$r$ MoE. A single null logit suffices since all copies are identical.

\subsection{Controlling Data Sparsity}

For $N$ real experts and target data sparsity $\rho \in (0, 1]$, we set:
\begin{equation}
    M = N \cdot \frac{1 - \rho}{\rho}
\end{equation}
With $N = 64$: $\rho = 0.5$ requires $M = 64$ null copies; $\rho = 0.25$ requires $M = 192$.

\subsection{Training Objectives}

\paragraph{Load Balancing Loss.} We apply a global load balancing \cite{qiu2025demonsdetailimplementingload} over all $N + M$ slots:
\begin{equation}
    \mathcal{L}_{\text{bal}} = \hldelta{(N + M)} \cdot \sum_{i=1}^{\hldelta{N+M}} f_i \cdot P_i
\end{equation}
where $f_i$ is the fraction of tokens routed to slot $i$ and $P_i$ is the average routing probability. Enforcing uniform load across an expanded pool that includes $M$ null copies incentivizes the target data sparsity.

\paragraph{Z-Loss.} We apply z-loss to stabilize training \cite{chowdhery2022palmscalinglanguagemodeling, zoph2022stmoedesigningstabletransferable}:
\begin{equation}
    \mathcal{L}_z = \frac{1}{T} \sum_{t=1}^{T} \log^2 \left( \sum_{i=1}^{N+M} \exp(\tilde{G}(\bm{x}_t)_i) \right)
\end{equation}
With null experts, load balancing and task loss conflict more than in standard MoE: the model must route low-information tokens to nulls while achieving good performance on high-information tokens. Z-loss prevents the router from becoming overconfident.

\subsection{Properties}

\paragraph{Solution Space Preservation.} Higher-sparsity configurations can recover lower-sparsity solutions. Consider $\mathbf{K}^{4}_{0.5}$ versus $\mathbf{K}^{2}_{1.0}$: both have $\mathbb{E}[K] = 2$. If the optimal solution uses exactly 2 experts per token, the $\mathbf{K}^{4}_{0.5}$ router can learn to always select 2 real experts plus 2 null experts, exactly recovering the $\mathbf{K}^{2}_{1.0}$ output after renormalization. Data sparsity should only help: the model can always fall back to denser computation if sparsity does not benefit the data.

\paragraph{Soft Constraint.} Expert-choice \cite{zhou2022mixtureofexpertsexpertchoicerouting} enforces data sparsity as a hard constraint: experts have fixed capacity. Token-choice with nulls enforces it as a soft constraint: load balancing encourages but does not strictly enforce the target sparsity. The model learns to route low-information tokens to null experts as an emergent property of training.

\paragraph{Why Zero (Not Copy) Experts.} MoE++ \cite{jin2024moeacceleratingmixtureofexpertsmethods} explores several null expert variants: zero experts (output $= \bm{0}$), copy experts (output $= \bm{x}$), and constant experts. We use zero experts because only they preserve solution space. With copy experts at sparsity $\rho$, the output becomes approximately $(1 - \rho) \cdot \bm{x} + \rho \cdot \bm{y}_{\text{dense}}$—the input dominates regardless of what real experts compute, and the dense solution is no longer recoverable. See Appendix~A for empirical analysis.

\section{Experiments}
\subsection{Training Details}




We train on a multimodal mixture of vision-language data. Following MoMa~\cite{lin2024momaefficientearlyfusionpretraining}, we begin with a dense warmup phase (20k steps) before enabling MoE routing, allowing the model to develop meaningful representations before routing decisions. After warmup, we upcycle to MoE using a fine-grained expert shape inspired by~\cite{dai2024deepseekmoeultimateexpertspecialization}: 64 experts at $4\times$ granularity, with 30\% of FFN parameters reinitialized randomly. We upcycle from 0.6B and 1.7B Qwen3~\cite{yang2025qwen3technicalreport} dense models, yielding 5.3B-A1.2B and 18.7B-A3.4B model scales at $\mathbb{E}[K] = 4$.

For optimization, we use AdamW~\cite{loshchilov2019decoupledweightdecayregularization} with a peak learning rate of 2e-5 (tuned across all experiments), $\beta = (0.9, 0.95)$, and weight decay of 0.1. We apply a WSD~\cite{wen2024understandingwarmupstabledecaylearningrates} learning rate schedule with 500-step warmup, decaying to 10\% of peak over the final 10\% of training. Load balancing and z-loss weights are set to 2e-2 and 1e-3 respectively.

In Section~\ref{sec:scaling_data_sparsity} and Section~\ref{sec:compute_efficiency} we train for 50k steps with batch size 512 and sequence length 2048 ($\sim$52B tokens total). In Section~\ref{sec:hero_run} we extend training to 200k steps with batch size 128 and sequence length 8192 ($\sim$209B tokens total). Remaining details can be found in~\cite{Isaac0.1}.

\subsection{Scaling Data Sparsity}
\label{sec:scaling_data_sparsity}
\FigureScalingSparsity

We first investigate how performance varies across data sparsity levels at fixed expected compute. We train 9 configurations at $\mathbb{E}[K] = 2$, varying base scale and sparsity:

\begin{itemize}[leftmargin=1.2em]
    \item 0.6B: $\mathbf{K}^{2}_{1.0}$, $\mathbf{K}^{3}_{0.67}$, $\mathbf{K}^{4}_{0.5}$, $\mathbf{K}^{8}_{0.25}$, $\mathbf{K}^{12}_{0.17}$
    \item 1.7B: $\mathbf{K}^{2}_{1.0}$, $\mathbf{K}^{3}_{0.67}$, $\mathbf{K}^{4}_{0.5}$, $\mathbf{K}^{8}_{0.25}$
\end{itemize}

We measure final training loss (averaged over the last 1k steps) and average eval performance across 10 standard benchmarks (AI2D~\cite{kembhavi2016ai2d}, A-OKVQA~\cite{schwenk2022aokvqa}, BLINK~\cite{fu2024blink}, ChartQA~\cite{masry2022chartqa}, Perceptron Grounding~\cite{Isaac0.1}, DocVQA~\cite{mathew2021docvqa}, M3Exam~\cite{zhang2023m3exam}, SEED-Bench~\cite{li2023seedbench}, TextVQA~\cite{singh2019textvqa}, VSR~\cite{liu2023vsr}). Results are shown in Figure~\ref{fig:scaling_sparsity}. Three findings emerge.

\paragraph{Solution space validation.} Data sparsity monotonically improves training loss across both model scales. This confirms solution space preservation: higher-sparsity configurations can recover lower-sparsity solutions, so performance is bounded below by the dense baseline.

\paragraph{Eval-loss divergence.} Eval gains do not track loss improvements at high sparsity. Performance peaks at $\rho \approx 0.5$, then degrades despite continued loss reduction. At $\mathbf{K}^{8}_{0.25}$, evals fall below baseline. We discuss possible explanations in Appendix~\ref{app:null_expert_limitations}.

\paragraph{Scale transfer.} The eval-optimal sparsity region ($\rho \in [0.5, 0.67]$) is consistent across 0.6B and 1.7B base scales. Sparsity configuration can be tuned at smaller scales and transferred, reducing experimental cost.

We constrain subsequent experiments to $\rho = 0.5$, where eval gains are stable. Extending the stable regime to higher sparsity through alternative routing mechanisms remains future work.

\subsection{Compute Efficiency Gains}
\label{sec:compute_efficiency}

Having established a stable operating regime at $\geq$50\% data sparsity, we now ask whether data sparsity improves compute efficiency across compute scales. We train models at two base sizes across multiple top-K values. For each (base model, top-K) pair, we compare three data sparsities: the 1.0 dense baseline, 0.67, and 0.5. Results are shown in Figure~\ref{fig:compute_efficiency}.

\FigureCompEff

Data sparsity reveals a more compute-efficient frontier in both training loss and evals, with gains more pronounced at larger compute scales. The effect is clearest in training loss, where data-sparse configurations consistently outperform dense baselines at matched expected FLOPs. Eval gains are present but slightly noisier.

\subsection{Hero Run}
\label{sec:hero_run}

To confirm these gains hold at larger compute budgets, we extend training of the 1.7B $\mathbf{K}^{4}_{1.0}$ and $\mathbf{K}^{8}_{0.5}$ models to 200k steps (~209B tokens). Table~\ref{tab:results-general} compares our models against Isaac 0.2, a dense 1.7B baseline trained with the same recipe, and InternVL3.5-20B-A4B \cite{wang2025internvl35advancingopensourcemultimodal}, a similarly sized sparse VLM. Note that InternVL3.5 was trained with more compute across multiple stages, whereas our models use only single-stage SFT; we report their published results for reference. Both MoE configurations consistently outperform the dense baseline across all task categories, with the $\mathbf{K}^{8}_{0.5}$ model achieving the best overall results. The improvements are particularly pronounced in OCR and counting tasks. Despite using a simpler training pipeline, our models remain competitive with InternVL and surpass it on several benchmarks.

\providecommand{\tablehead}[1]{\textbf{#1}}
\providecommand{\papersym}{\textsuperscript{\dag}}

\begin{table}[h]
\centering
\tiny
\begin{threeparttable}
\renewcommand{\arraystretch}{1.2}
\setlength{\heavyrulewidth}{1.0pt}
\setlength{\lightrulewidth}{0.6pt}
\setlength{\cmidrulewidth}{0.6pt}
\setlength{\cmidrulesep}{0.25em}

\newcommand{\mytoprule}{\toprule}
\newcommand{\mymidrule}{\midrule}
\newcommand{\mybottomrule}{\bottomrule}
\newcommand{\mycmidrulecolumntwo}{\addlinespace[.2ex]\cmidrule(lr){2-6}\addlinespace[.1ex]}
\newcommand{\mysectionrule}{\addlinespace[.3ex]\cmidrule[\heavyrulewidth](lr){1-6}\addlinespace[.3ex]}
\begingroup
\arrayrulecolor{black}
\setlength{\heavyrulewidth}{1.0pt}
\setlength{\lightrulewidth}{0.6pt}
\setlength{\cmidrulewidth}{0.6pt}
\setlength{\cmidrulesep}{0.25em}

\begin{adjustbox}{max width=\textwidth,keepaspectratio}
\begingroup\let\origscalebox\scalebox\renewcommand{\scalebox}[2]{#2}

\newcommand{\tblcell}{\footnotesize}
\newcolumntype{L}{>{\tblcell}l}
\newcolumntype{C}{>{\tblcell}c}

\begin{tabular}{L L C C C C}
\mytoprule
\textbf{Task} & \textbf{Benchmark} & \tablehead{1.7B $\mathbf{K}^{4}_{1.0}$} & \tablehead{1.7B $\mathbf{K}^{8}_{0.5}$} & \tablehead{Isaac 0.2} & \tablehead{InternVL3.5-20B-A4B} \\
\mymidrule
Size & -- & 18B-A3B & 18B-A3B & 2B & 20B--A4B \\
\mymidrule

\multirow[t]{14}{*}{Pointing}
& Aerial Grounding & 80.7 & \textbf{82.2} & 73.1 & -- \\
& Perceptron Grounding & 58.8 & \textbf{59.2} & 51.5 & -- \\
& RefCOCO \cite{yu2016refcoco} & 87.6 & 87.8 & 87.4 & \textbf{91.9} \\
\mycmidrulecolumntwo
& Overall & 75.7 & \textbf{76.4} & 70.7 & -- \\

\mymidrule
\multirow[t]{5}{*}{OCR}
& ChartQA \cite{masry2022chartqa} & 79.0 & 80.3 & 75.1 & \textbf{86.6} \\
& DocVQA \cite{mathew2021docvqa} & 92.9 & \textbf{93.8} & 92.1 & 92.9  \\
& A\mbox{-}OKVQA (val) \cite{schwenk2022aokvqa} & 89.4 & \textbf{91.8} & 87.2 & -- \\
& TextVQA \cite{singh2019textvqa} & 80.8 & \textbf{82.0} & 78.1 & 78.5 \\
& OCRBench \cite{Liu_2024} & 873 & \textbf{880} & 857 & 870  \\
\mycmidrulecolumntwo
& Overall & 85.9 & \textbf{87.2} & 83.6 & -- \\

\mysectionrule

\multirow[t]{10}{*}{Counting}
& Aerial Counting & 53.0 & \textbf{57.0} & 52.0 & -- \\
& CVBench \cite{zhu2026cvbenchbenchmarkingcrossvideosynergies} & 72.1 & \textbf{73.8} & 72.5 & -- \\
& PixMoCount \cite{pixmocount_2025} & 68.6 & \textbf{69.4} & 66.7 & -- \\
& CountBench \cite{paiss2023teachingclip} & 85.5 & \textbf{87.5} & 84.6 & -- \\
\mycmidrulecolumntwo
& Overall & 69.8 & \textbf{71.9} & 69.0 & -- \\

\mymidrule
\multirow[t]{21}{*}{General}
& VSR (Zero\mbox{-}Shot) \cite{liu2023vsr} & 79.6 & \textbf{80.6} & 78.6 & -- \\
& VQA v2 \cite{goyal2017makingVQA} & \textbf{82.6} & 82.4 & 80.8 & -- \\
& RealWorldQA \cite{xai2024realworldqa} & 74.1 & 75.1 & \textbf{77.9} & 71.2 \\
& SEED-Bench \cite{li2023seedbench} & \textbf{76.8} & 75.8 & 74.2 & -- \\
& M3Exam (English) \cite{zhang2023m3exam} & 54.8 & \textbf{58.6} & 55.8 & -- \\
& NLVR2 \cite{suhr2019nlvr2} & 82.6 & \textbf{83.2} & 76.4 & -- \\
& BLINK \cite{fu2024blink} & 53.8 & 55.6 & 55.0 & \textbf{59.0} \\
& MathVista \cite{lu2024mathvista} & 73.2 & 73.9 & 69.6 & \textbf{78.0} \\
& MME \cite{fu2023mme} & 2221 & 2237 & 2092 & \textbf{2318} \\
& AI2D \cite{kembhavi2016ai2d} & 79.1 & 78.9 & 74.5 & \textbf{85.9} \\
& ERQA \cite{erqa2025} & 40.6 & 39.4 & 36.8 & \textbf{41.6} \\
\mycmidrulecolumntwo
& Overall & 70.6 & \textbf{71.2} & 68.6 & -- \\

\mybottomrule
\end{tabular}

\endgroup
\end{adjustbox}
\endgroup

\vspace{1mm}
\caption{Benchmark performance of extended training runs (200k steps, 209B tokens). Our sparse MoE models (18B-A3B) consistently outperform the dense Isaac 0.2 baseline across pointing, OCR, counting, and general vision-language tasks, while remaining competitive with InternVL3.5-20B-A4B despite its more complex multi-stage training pipeline.}
\label{tab:results-general}
\end{threeparttable}
\end{table}

\section{Routing Behavior}
The previous sections established that data sparsity improves the compute-efficiency frontier. But how does the model use this flexibility? If tokens route to null experts uniformly at random, data sparsity would offer no advantage. The value comes from selective allocation: routing low-information tokens to null experts while preserving compute for tokens that need it.

We find the model learns exactly this. Three patterns emerge from routing analysis. First, null experts shift compute from vision to text without explicit modality routing. Second, this reallocation is task-dependent: the same image receives different compute maps under different prompts. Third, we measure increasing polarization in MoE compute per token at high data sparsity: tokens route entirely to real or null experts rather than mixing.

\subsection{Modality Compute Rebalancing}

\FigureModalityCompute
Null experts invert the compute distribution between modalities. To quantify this, we measure three quantities: token distribution (each modality's share of total tokens, constant at 78\% vision / 22\% text across configurations), compute distribution (each modality's share of total effective compute, where a token routed entirely to null experts contributes 0 and one using only real experts contributes 1), and compute intensity (the average compute score per token, indicating what fraction of expert slots are filled by real versus null experts).

Figure~\ref{fig:modality_compute} shows these metrics under varying data sparsity. In dense configurations without null experts ($\mathbf{K}^{2}_{1.0}$), compute distribution mirrors token distribution: vision consumes 78\% of MoE compute simply because it produces 78\% of tokens. This is inefficient. Vision encoders generate many redundant patches—blank regions, repetitive textures, uninformative background—while text tokens tend to be information-dense.

As we introduce null experts with increasing sparsity ($\mathbf{K}^{3}_{0.67}$ through $\mathbf{K}^{12}_{0.17}$), vision tokens route to null experts far more aggressively than text. At $\mathbf{K}^{12}_{0.17}$, vision compute intensity drops to 4\% compared to 22\% for text, producing an inverted compute distribution: text consumes 60\% of total compute despite representing only 22\% of tokens. The model allocates compute based on information content rather than token count.

Prior work has noted that modality imbalance poses challenges for multimodal MoE training. V-MoE~\cite{riquelme2021scalingvisionsparsemixture} addresses heterogeneous token importance via Batch Prioritized Routing: tokens are ranked by an importance score and, under capacity constraints, the least informative patches are dropped. This enables adaptive per-image compute but requires explicitly scoring and sorting tokens across the batch. LiMoE~\cite{mustafa2022multimodalcontrastivelearninglimoe} tackles training instability from modality imbalance through entropy-based auxiliary losses applied per-modality, finding that modality-specific expert specialization emerges organically with proper regularization. MoMa~\cite{lin2024momaefficientearlyfusionpretraining} takes an architectural approach, partitioning experts into modality-specific groups where each group exclusively processes its designated modality—text tokens route only to text experts, image tokens only to image experts. Using the null expert mechanism we achieve similar modality-aware rebalancing implicitly: standard load balancing over an expanded routing pool is sufficient to induce differential compute allocation without explicit importance scoring, per-modality regularization, or architectural partitioning. The model receives no signal about which modality deserves more compute; it learns this from task performance alone.

\subsection{Token-level Compute Maps}

\FigureComputeMap

Figure~\ref{fig:compute_map_0} visualizes per-token compute utilization. For each token, we measure the fraction of top-K slots filled by real experts (averaged across MoE layers), producing a score between 0 (complete null routing) and 1 (full computation). 

Compute varies substantially within each modality, not just between them. Among vision tokens, salient regions receive more compute than background. Among text tokens, predictable continuations, punctuation, and control sequences route to null experts more aggressively than information-dense tokens. The model allocates compute based on token-level information content, not modality alone. This within-modality variation implies data sparsity benefits text-only training as well-we focus on multimodal settings because vision token heterogeneity is pronounced, but text sequences contain their own redundancy, and the model exploits it. Prior work confirms this~\cite{meituanlongcatteam2025longcatflashtechnicalreport, jin2024moeacceleratingmixtureofexpertsmethods, zeng2024adamoetokenadaptiveroutingnull}. See Appendix~\ref{app:compute_maps} for more token-level maps.

\FigureSysPromptCompute

Allocation is also context-dependent. The same image receives different compute maps under different prompts. Figure~\ref{fig:sys_prompt_compute} compares routing for identical visual input with a underspecified prompt (i.e., ``Answer with single word.'') versus a targeted segmentation prompt. Under the underspecified task, the model distributes compute broadly. Under segmentation, it concentrates compute on task-relevant regions and routes most patches to null experts. Null expert routing is not a fixed property of inputs-the model learns that "relevant" is task-relative and skips computation for tokens that do not serve the current objective.

\subsection{Data Sparsity Strategies}

\FigureDataSparsityStrategy

How do models achieve data sparsity? They could spread null routing uniformly—every token uses slightly fewer real experts—or polarize, with some tokens using full compute and others using none. Figure~\ref{fig:null_expert_utilization} shows the distribution of real expert utilization across configurations.

At fixed $\mathbb{E}[K]=2$, increasing data sparsity (decreasing $\rho$) produces an almost linear increase in the fraction of tokens routed to zero real experts. The $\mathbf{K}^{12}_{0.17}$ configuration routes over 60\% of tokens to zero real experts despite having 12 slots available—most selections fall on null copies. The model makes binary decisions: full computation or none.

Increasing $\mathbb{E}[K]$ reduces this polarization. Models with higher expected active experts distribute compute more uniformly across tokens rather than concentrating it on a subset.
Scaling base dense model size has a similar marginal effect: 1.7B models show lower zero-compute ratios than 0.6B at equivalent configurations, though the effect is modest compared to the sparsity target itself.

\section{Conclusion and Future Work}

Weight and data sparsity are orthogonal efficiency axes, but standard MoE exploits only the first: each token activates a fixed subset of experts regardless of its information content. Composing both yields a strictly better compute-efficient frontier, and null experts achieve this composition within token-choice MoE while preserving causality.

We demonstrated these gains on vision-language model training, where data heterogeneity is pronounced. At matched expected FLOPs, data-sparse configurations achieved lower training loss and improved downstream performance across model scales. The model learned to allocate compute by information content rather than token count: vision tokens routed to null experts more aggressively than text, and routing patterns shifted based on task demands, all without explicit supervision. These behaviors emerged from the standard load balancing objective alone.

Our framing of data sparsity as an axis orthogonal to weight sparsity opens several directions. Null experts are one mechanism for achieving data sparsity, but not necessarily the optimal one -- the eval breakdown at high sparsity Section~\ref{sec:scaling_data_sparsity}) suggests limitations in jointly modeling both axes through a single softmax, and understanding this failure mode could inform better architectures. More elaborate router distribution shaping approaches \cite{mirvakhabova2025dirichletpriorshapingguidingexpert} offer a promising alternative: rather than encouraging data sparsity through load balancing over an expanded pool, one could directly fit the global routing distribution to a target encoding balanced load, sparsity level, and other desiderata simultaneously. Data sparsity also extends beyond MoE layers to attention modules, where some tokens may warrant less cross token compute than others \cite{jin2025mohmultiheadattentionmixtureofhead}.


\bibliographystyle{unsrtnat}
\bibliography{references}

\appendix
\clearpage

\section{Null Experts}

\subsection{Inference Limitations}
\label{app:null_expert_limitations}

In Figure~\ref{fig:scaling_sparsity}, training loss improves monotonically as we increase data sparsity, but evaluation peaks around $\rho \approx 0.5$ and degrades at more aggressive sparsity (e.g., $\mathbf{K}^{8}_{0.25}$, $\mathbf{K}^{12}_{0.17}$). We view this gap as a limitation of the \emph{single-softmax, thresholded null-copy} construction rather than evidence that data sparsity is inherently harmful. Below we outline three interacting effects that become more pronounced as $\rho$ decreases.

\paragraph{1) Effective router resolution collapses at high sparsity.}
With $M$ duplicated null copies, the router produces a categorical distribution over $N{+}M$ slots, and real experts must compete against a large null block in the same softmax. When $\rho \ll 1$, the balancing objective encourages substantial probability mass to sit on the null region of the simplex. This can reduce the \emph{effective} resolution of the router among real experts: relative differences between real experts are compressed after normalization, and gradients that refine expert identity decisions can be attenuated because most mass lies outside the real-expert subset. In this regime the router can remain good at the \emph{compute} decision (real vs.\ null) while becoming worse at the \emph{expert identity} decision (which real expert), plausibly harming downstream metrics even as likelihood improves.

\paragraph{2) Thresholding becomes unstable and encourages polarization.}
Null-copy routing implements a top-K threshold: a token activates $r$ real experts when exactly $r$ real logits exceed the null logit. At aggressive sparsity, the null threshold must sit above most real logits, making intermediate allocations (e.g., $r=1,2,3$) sensitive to small logit perturbations. A robust alternative for the model is to adopt a bimodal strategy: push the null logit far above all real logits (route to all nulls) or far below many of them (route to mostly real experts). Such polarization can hurt evaluation when tasks benefit from \emph{broad but shallow} compute (many moderately informative tokens) rather than concentrating compute on a small subset.

\paragraph{3) Auxiliary-objective interference grows as $\rho$ decreases.}
Standard MoE load balancing assumes every routed slot corresponds to meaningful computation. With null copies, the balancing term implicitly asks the router to distribute tokens uniformly over both (i) semantically specialized experts and (ii) a replicated ``do nothing'' region. At high sparsity, satisfying balancing can become easier by increasing null usage rather than improving expert specialization, increasing the mismatch between what balancing incentivizes and what evaluation rewards. This interference is amplified by renormalization: once a token selects any real experts, we renormalize over them, so the forward pass is insensitive to how much probability mass the router placed on nulls \emph{as long as the top-$k$ set is unchanged}. In contrast, the auxiliary losses depend on the full softmax distribution, enabling progress on auxiliary objectives that does not necessarily translate into better expert assignments.

\paragraph{Empirical signatures.}
These mechanisms make predictions that are consistent with the routing behavior we already observe. As $\rho$ decreases, routing becomes increasingly polarized (Figure~\ref{fig:null_expert_utilization}): a growing fraction of tokens route to zero real experts, while the remainder consume most of the compute budget. This is the expected qualitative signature of an unstable thresholding regime, where intermediate allocations are fragile and the model prefers an all-or-nothing strategy. In the same regime, reduced discrimination among real experts becomes more likely because real experts must compete against a large null block in the softmax; this provides a plausible explanation for why likelihood can continue to improve while downstream metrics degrade.

\paragraph{Implication.}
Taken together, these effects suggest that the main limitation is \emph{coupling} the compute decision (how many real experts) and the expert identity decision (which real experts) inside a single normalized routing distribution whose support is dominated by null copies at low $\rho$. Extending stable evaluation gains to more aggressive data sparsity likely requires modifying this coupling or the associated regularization; we leave this to future work and restrict subsequent experiments to $\rho \ge 0.5$, where evaluation gains are stable.

\subsection{Copy Experts}

We initially explored copy experts following prior work, but abandoned this direction due to both theoretical and empirical concerns.

\paragraph{Solution space violation.} Copy experts break the solution space preservation property that makes null experts attractive. With copy experts at data sparsity $\rho$, the MoE output becomes approximately $(1 - \rho) \cdot \bm{x} + \rho \cdot \bm{y}_{\text{dense}}$, where the input $\bm{x}$ dominates regardless of what real experts compute. The dense baseline ($\mathbf{K}^{k}_{1.0}$) is therefore not recoverable within the data-sparse configuration. This residual dilution effect worsens at higher sparsity—at extreme settings, the MLP output collapses toward the identity function. We suspect this explains why LongCat-Flash investigated only modest data sparsity (75\%).

\paragraph{Polarized routing.} The residual dilution problem incentivizes polarized routing behavior. When a token routes to a mixture of real and copy experts, the output is diluted by the copy expert contributions. The model can avoid this dilution by making binary decisions: route entirely to real experts (preserving full expert output) or entirely to copy experts (clean residual connection). Figure~\ref{fig:compute_map_copy_expert} shows this effect in token-level compute maps from copy expert runs, which exhibit far more polarized compute distributions than their zero expert counterparts. With zero experts, this pressure does not exist—routing weights are renormalized over only the selected real experts, so mixed routing incurs no penalty.

\paragraph{Pathological training dynamics.} The combination of residual connections and load balancing loss creates a strong gradient pathway through copy experts. In some cases, the model learned to rely on copy expert routing as a default strategy rather than using null routing selectively for low-information tokens. This produced pathological behaviors: the model learned inverse mappings that allowed strong SigLIP embeddings to pass through unchanged, and in early training (within the first few hundred steps), learned to drop virtually all vision tokens to satisfy sparsity targets while relying solely on attention and the vision encoder for visual conditioning.

\paragraph{Mitigation.} For practitioners who wish to use copy experts despite these issues, we found that both dense warmup and null expert warmup are essential to prevent the model from dropping all vision tokens early into training but compute maps remained polarized.

\FigureComputeMapCopyExpert

\section{Infrastructure}

Our implementation of null experts is a minimal extension of standard token-choice MoE, adding minimal lines of code to the routing logic. This simplicity is possible because token-choice MoE already contains a natural pocket of dynamic computation that we can exploit.

Many early MoE implementations restored static shapes by imposing expert capacity-GShard-style routing~\citep{lepikhin2020gshardscalinggiantmodels}, for instance, drops or truncate tokens when experts overflow. Others relied on specialized sparse kernels like MegaBlocks~\citep{gale2022megablocksefficientsparsetraining} to remain dropless while handling variable loads. More recently, implementations have converged on grouped GEMM: tokens are permuted into contiguous per-expert blocks, a single grouped matrix multiplication executes over the resulting operands, and outputs scatter back to the original token order. PyTorch~\cite{pytorch} 2.8's native \texttt{GroupedGEMM} handles highly variable tokens-per-expert efficiently with only minimal alignment padding.

This is what makes null experts essentially free. The kernel already accepts variable token counts, so we simply expand the router to include null logits and truncate the sorted token list before the same \texttt{grouped\_mm} call. The kernel itself is unchanged. Algorithm~\ref{alg:null-moe} shows the implementation in PyTorch-style pseudocode, with \hldeltabox{highlighted lines} denoting changes from standard token-choice. Because \texttt{argsort} places null expert indices ($\geq N$) at the end of the sorted order, we slice \texttt{[:num\_real]} and proceed as usual.

\paragraph{Asynchronous load-balancing loss.} Global load-balancing losses are critical for stable token-choice MoE training~\citep{qiu2025demonsdetailimplementingload} and remain our mechanism for controlling data sparsity. However, they require additional collectives in each MoE layer-typically all-reduces over per-expert token counts. To keep these off the forward critical path, we launch them asynchronously inside each layer, continue without waiting, and synchronize only after the forward completes to finalize the auxiliary loss.

\paragraph{Adaptive activation checkpointing.} With null experts, memory requirements shift during training. Early on, routers utilize the full configured top-K. As load balancing converges, the effective number of active experts drops for most tokens, and a checkpointing configuration tuned for the first ${\sim}100$ steps becomes overly conservative. We address this with an adaptive controller that periodically measures peak memory and headroom, disabling checkpointing when safe to increase throughput and re-enabling it if memory becomes constrained. Since policy updates occur only every $N$ steps, recompilation overhead is negligible.

\begin{algorithm}[t]
\caption{Token-Choice MoE with Null Experts}\label{alg:null-moe}
\begin{algorithmic}[1]
\Require \texttt{x}: $(T, D)$, \texttt{W}: $(\hldelta{N+1}, D)$, \texttt{E}: $(N, D_h, D)$, top-K, null copies $M$
\Statex
\LineComment{Router}
\State \texttt{logits = x @ W.T}  \Comment{$(T, N+1)$}
\State \hldeltabox{\texttt{logits = cat([logits[:, :N], logits[:, N:].expand(-1, M)], dim=1)}}  \Comment{$(T, N+M)$}
\State \texttt{scores = softmax(logits, dim=1)}
\State \texttt{top\_scores, top\_idx = top-K(scores, k, dim=1)}  \Comment{$(T, k)$}
\Statex
\LineComment{Reorder tokens by expert (standard token-choice)}
\State \texttt{order = argsort(top\_idx.flatten(), stable=True)}
\State \texttt{sorted\_scores = top\_scores.flatten()[order]}
\State \texttt{sorted\_tokens = order // k}
\State \texttt{num\_per\_expert = histc(top\_idx, bins=}\hldelta{\texttt{N+M}}\texttt{)}  \Comment{$(N+M,)$}
\Statex
\LineComment{\hldeltabox{Truncate to real experts (null experts sort last)}}
\State \hldeltabox{\texttt{num\_real = num\_per\_expert[:N].sum()}}
\State \hldeltabox{\texttt{sorted\_scores[:num\_real] /= sorted\_scores[:num\_real].sum\_per\_token()}}  \Comment{renorm}
\Statex
\LineComment{Expert computation (unchanged \texttt{grouped\_mm})}
\State \texttt{x\_in = x[sorted\_tokens}\hldelta{\texttt{[:num\_real]}}\texttt{] * sorted\_scores}\hldelta{\texttt{[:num\_real, None]}}
\State \texttt{x\_out = grouped\_mm(x\_in, E, num\_per\_expert}\hldelta{\texttt{[:N]}}\texttt{)}  \Comment{same kernel}
\Statex
\LineComment{Scatter back}
\State \texttt{y = zeros(T, D)}
\State \texttt{y.scatter\_add\_(0, sorted\_tokens}\hldelta{\texttt{[:num\_real]}}\texttt{, x\_out)}
\State \Return \texttt{y}
\end{algorithmic}
\label{algo_pseudocode}
\end{algorithm}

\section{Compute Maps}
\label{app:compute_maps}

\FigureComputeMapDocVQA

\FigureComputeMapLayers

\end{document}

%% file: figures.tex
\newcommand{\FigureCompEff}{
\begin{figure}[H]
  \centering
  \includegraphics[width=1.0\linewidth]{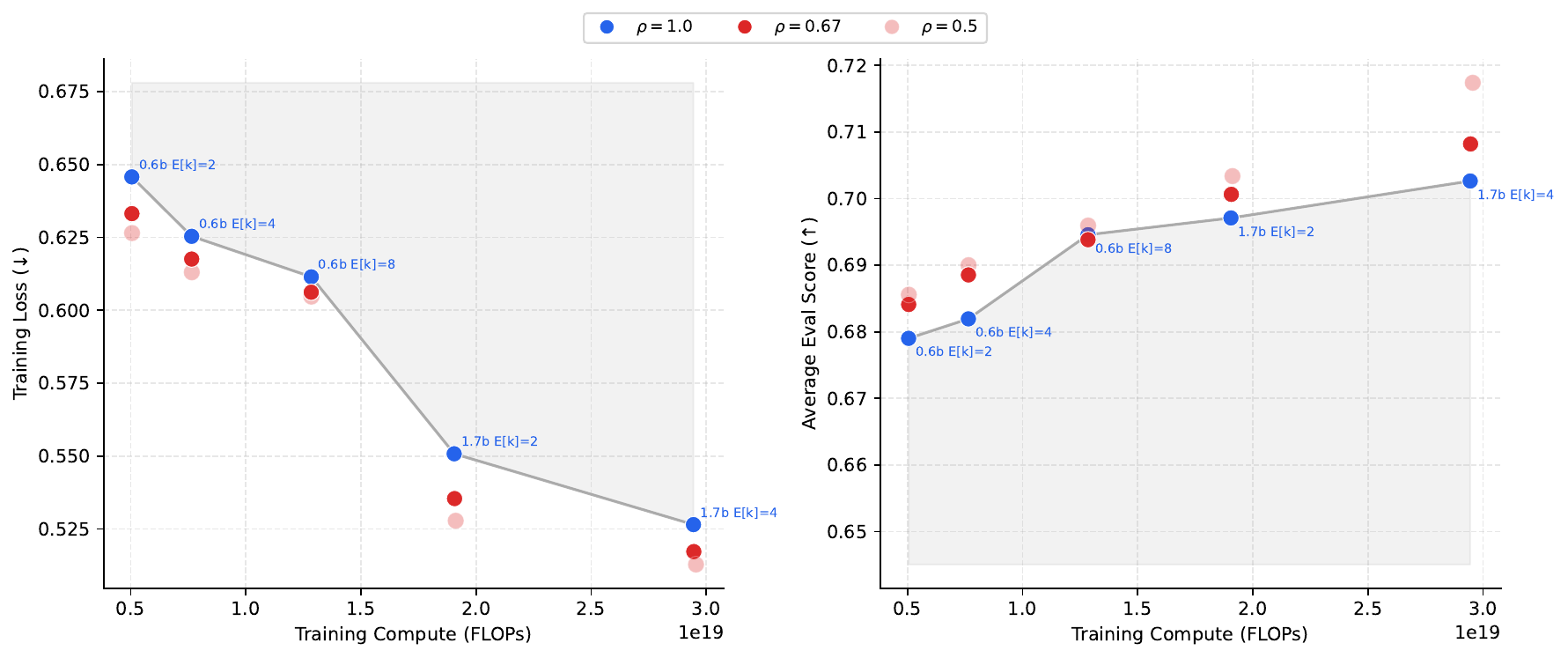}
  \caption{Training loss \textbf{(Left)} and average evaluation score \textbf{(Right)} as a function of training FLOPs for data-dense and data-sparse MoE models. Blue circles represent data-dense baselines with fixed top-K routing; red circles represent data-sparse models using null experts, with opacity indicating the sparsity ratio $\rho$. The gray line connects data-dense models to form a Pareto frontier, with the shaded region indicating suboptimal performance. Data-sparse models consistently outperform data-dense baselines at equivalent compute budgets, achieving lower training loss and higher evaluation scores. Labels indicate model scale and expected active experts $\mathbb{E}[K]$; $\rho$ denotes the ratio of real experts to total experts (e.g., $\rho=0.50$ means half of selected experts are real on average).}
  \label{fig:compute_efficiency}
\end{figure}
}

\newcommand{\FigureScalingSparsity}{
\begin{figure}[H]
  \centering
  \includegraphics[width=1.0\linewidth]{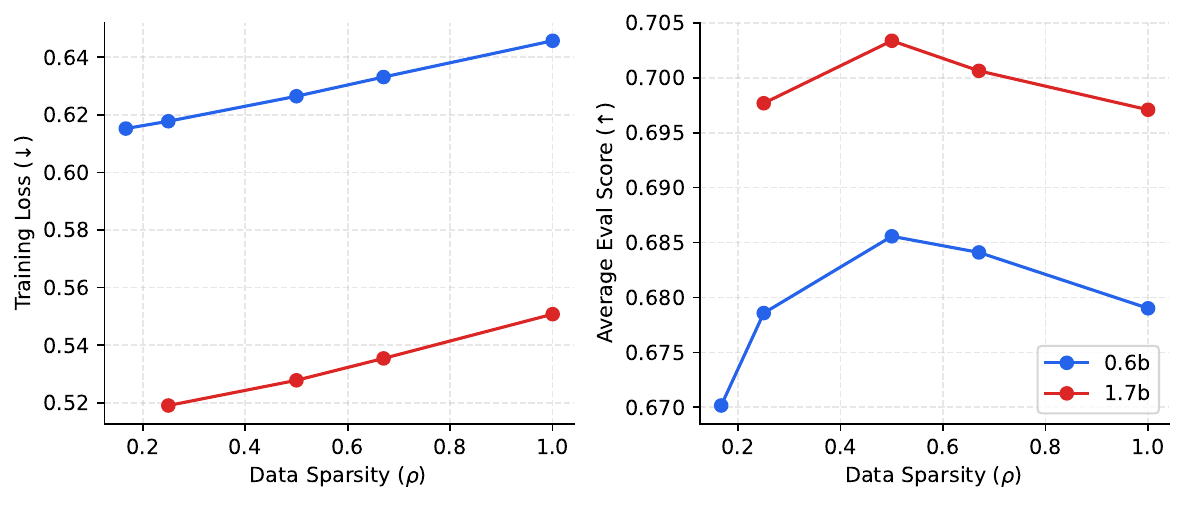}
  \caption{Effect of data sparsity ($\rho$) on training loss \textbf{(Left)} and evaluation score \textbf{(Right)} for MoE models upcycled from dense 0.6B and 1.7B models. Lower $\rho$ indicates more null experts. Training loss decreases monotonically with increased sparsity, while evaluation scores peak at $\rho \approx 0.5$.}
  \label{fig:scaling_sparsity}
\end{figure}
}

\newcommand{\FigureModalityCompute}{
\begin{figure}[H]
  \centering
  \includegraphics[width=1.0\linewidth]{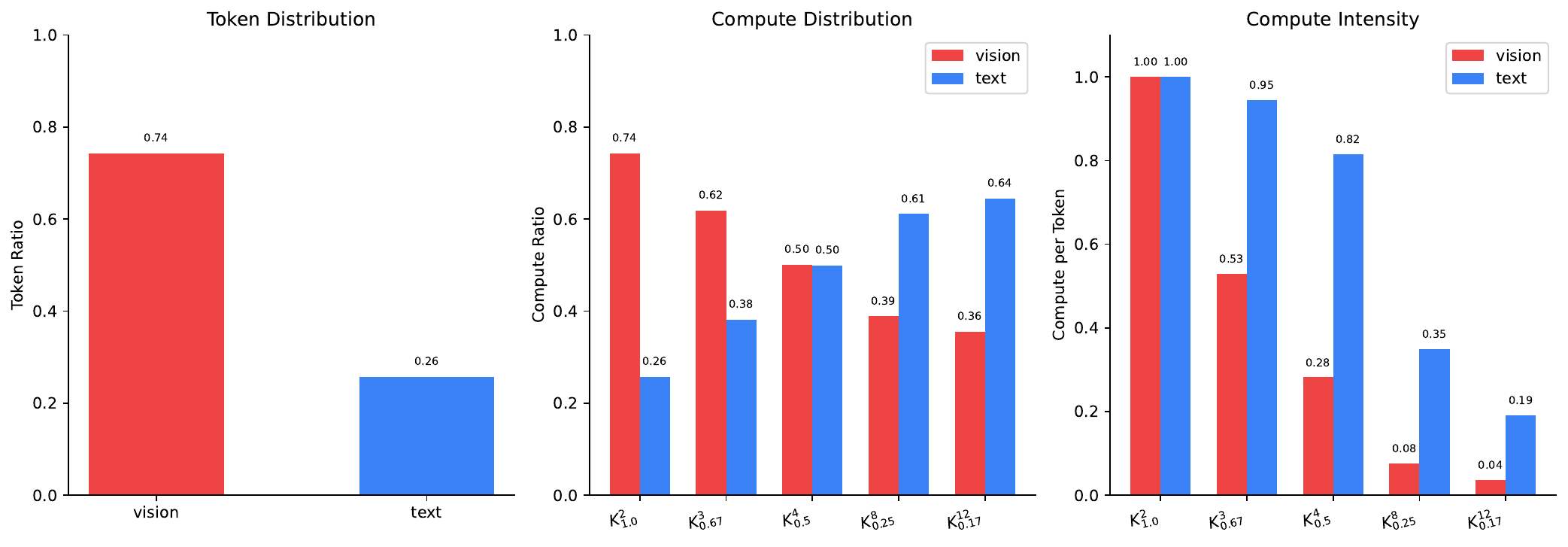}
  \caption{MoE compute distribution by modality across sparsity configurations. \textbf{Left:} Token distribution (constant at 74\% vision, 26\% text). \textbf{Middle:} Compute distribution. Dense configurations ($\mathbf{K}^{2}_{1.0}$) allocate compute proportionally to token count; data-sparse configurations route vision tokens to null experts, reducing vision's share from 74\% to 36\%. \textbf{Right:} Compute intensity (fraction of top-K slots filled by real experts). Vision intensity drops to 0.04 at $\mathbf{K}^{12}_{0.17}$ while text remains at 0.19.}
  \label{fig:modality_compute}
\end{figure}
}

\newcommand{\FigureDataSparsityStrategy}{
\begin{figure}[H]
  \centering
  \begin{subfigure}[b]{0.38\linewidth}
    \includegraphics[width=\linewidth]{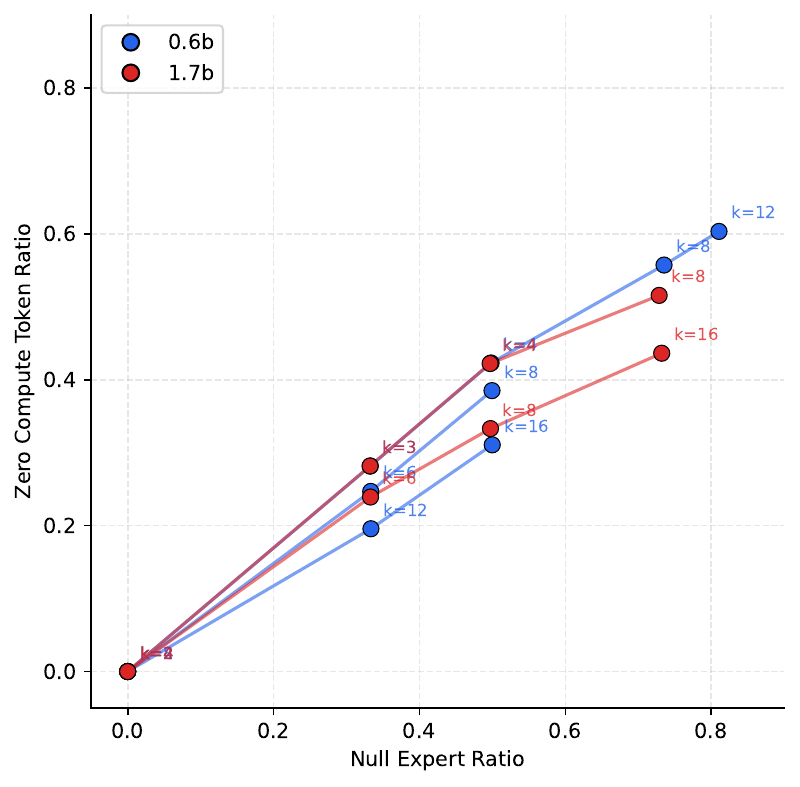}
    \caption{}
    \label{fig:null_expert_utilization:scatter}
  \end{subfigure}
  \hfill
  \begin{subfigure}[b]{0.29\linewidth}
    \includegraphics[width=\linewidth]{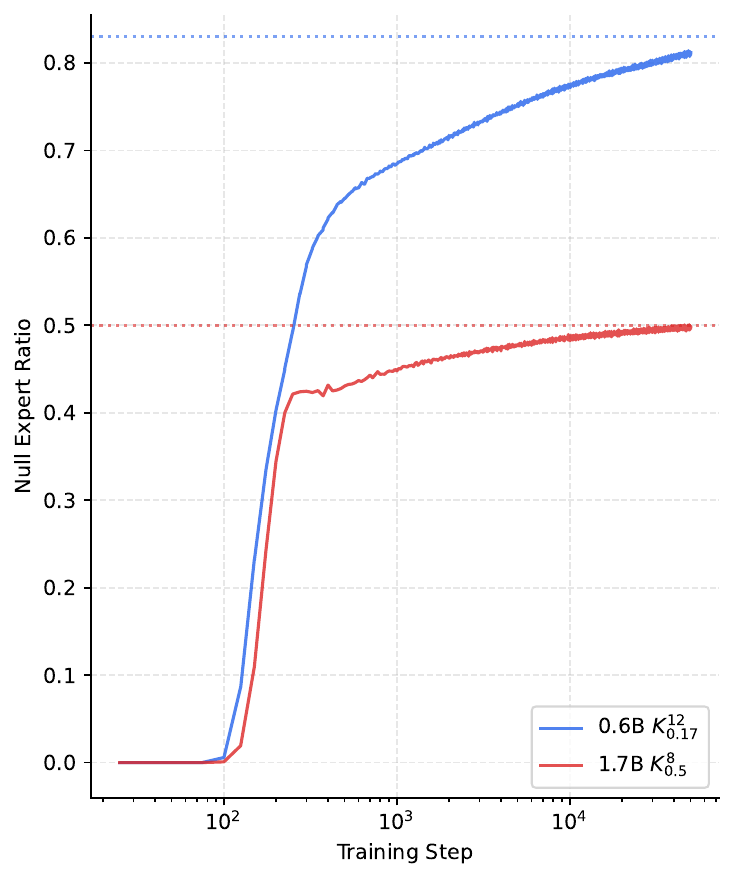}
    \caption{}
    \label{fig:null_expert_utilization:null_ratio}
  \end{subfigure}
  \hfill
  \begin{subfigure}[b]{0.29\linewidth}
    \includegraphics[width=\linewidth]{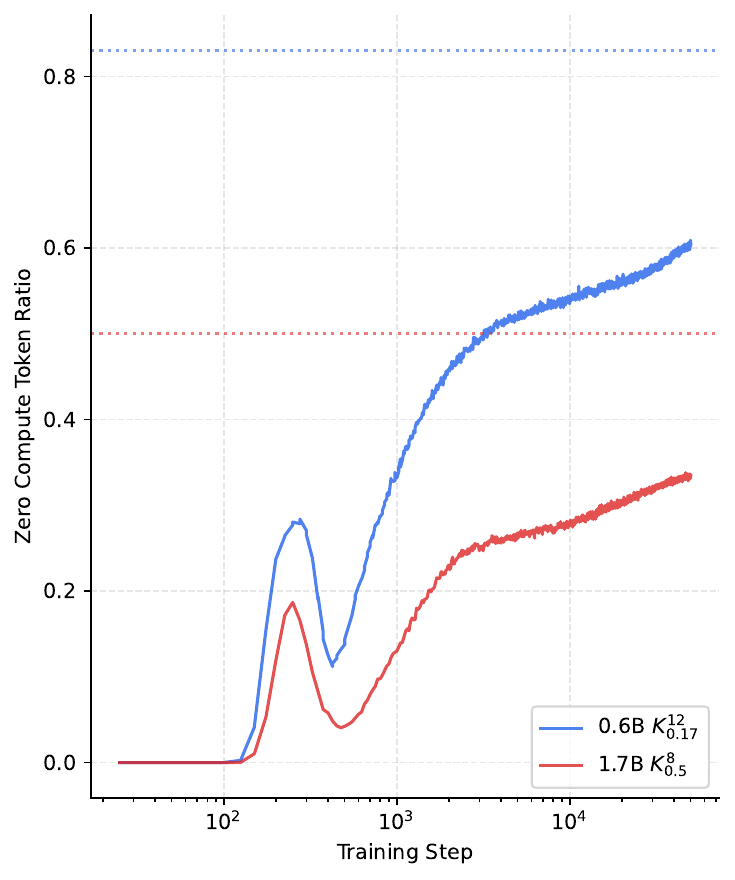}
    \caption{}
    \label{fig:null_expert_utilization:zero_compute}
  \end{subfigure}
  \caption{\textbf{(a)} Zero-compute token ratio versus null expert ratio, grouped by $\mathbb{E}[K]$ (labels show $k_{\text{max}}$). Configurations with the same $\mathbb{E}[K]$ achieve similar zero-compute ratios despite different $k_{\text{max}}$ values—expected active experts determines polarization, not $k_{\text{max}}$ alone. Zero-compute falls below null expert ratio because tokens not fully skipped can mix real and null experts within their top-K selection. \textbf{(b, c)} Training dynamics for 0.6B $\mathbf{K}^{12}_{0.17}$ and 1.7B $\mathbf{K}^{8}_{0.5}$. Both metrics converge toward theoretical sparsity (dotted); zero-compute consistently lags below null expert ratio.}
  \label{fig:null_expert_utilization}
\end{figure}
}

\newcommand{\FigureComputeMap}{
\begin{figure}[H]
  \centering
  \includegraphics[width=1.0\linewidth]{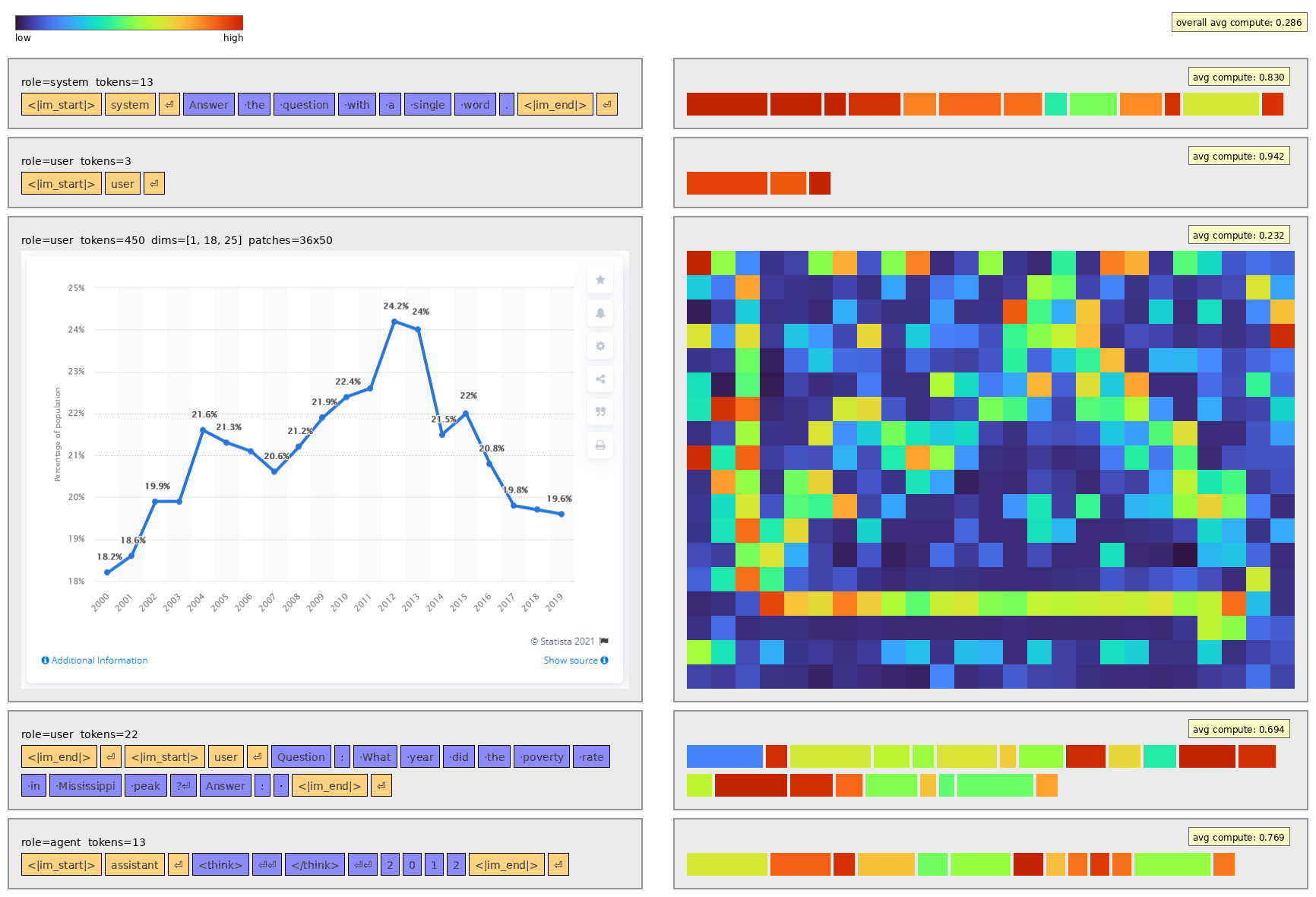}
  \caption{Per-token compute utilization for a sample sequence. \textbf{Left:} Original input (text and image patches). \textbf{Right:} Compute overlay where brightness indicates fraction of top-K slots filled by real experts—bright regions receive full computation, dark regions route to null experts. Inset values show mean compute per segment (image or text block) and overall sample compute (top-right). Vision tokens receive less compute on average due to redundant patches.}
  \label{fig:compute_map_0}
\end{figure}
}

\newcommand{\FigureSysPromptCompute}{
\begin{figure}[H]
  \centering
  \includegraphics[width=1.0\linewidth]{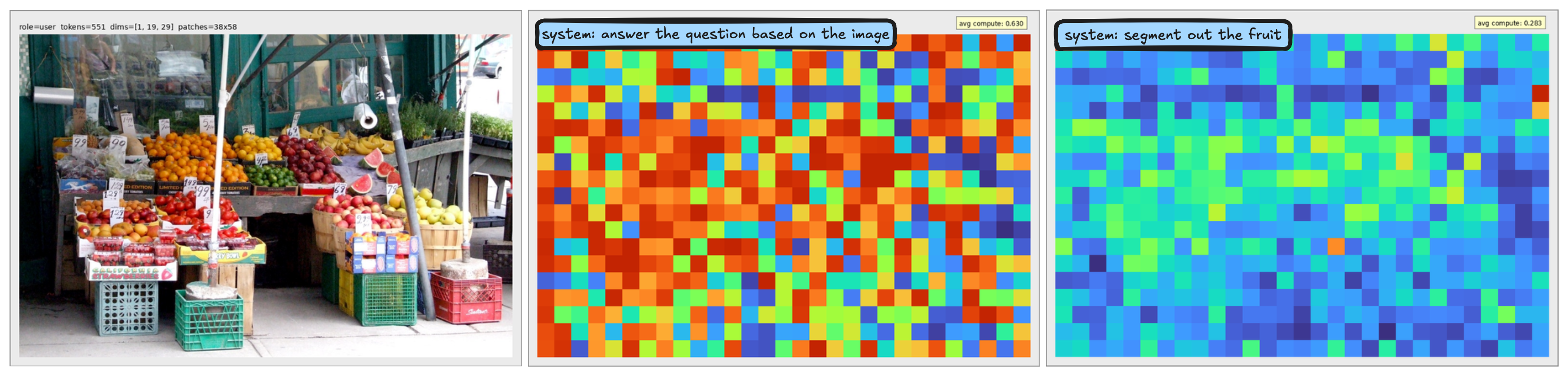}
  \caption{Compute allocation varies with task context. \textbf{Left:} Original input. \textbf{Middle:} Compute overlay under a general QA prompt—high compute distributed broadly. \textbf{Right:} Compute overlay under a targeted segmentation prompt—reduced overall compute, concentrated on task-relevant regions. System prompts shown top-left; mean compute statistics shown top-right. Brightness indicates real expert utilization (dark = null routing). The model routes more aggressively to null experts when the task requires only a subset of the image.}
  \label{fig:sys_prompt_compute}
\end{figure}
}

\newcommand{\FigureComputeMapCopyExpert}{
\begin{figure}[H]
  \centering
  \includegraphics[width=0.49\linewidth]{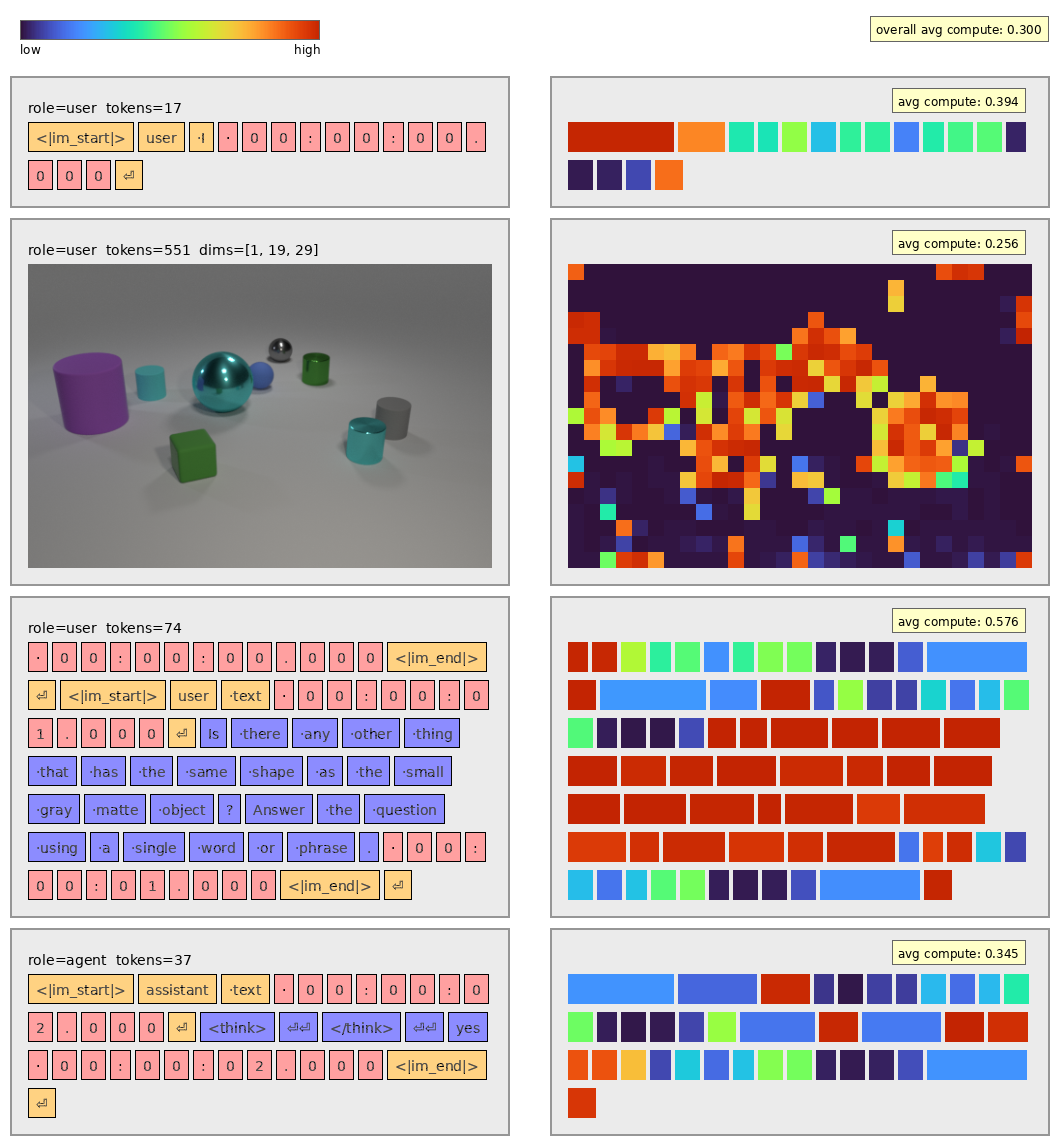}
  \hfill
  \includegraphics[width=0.49\linewidth]{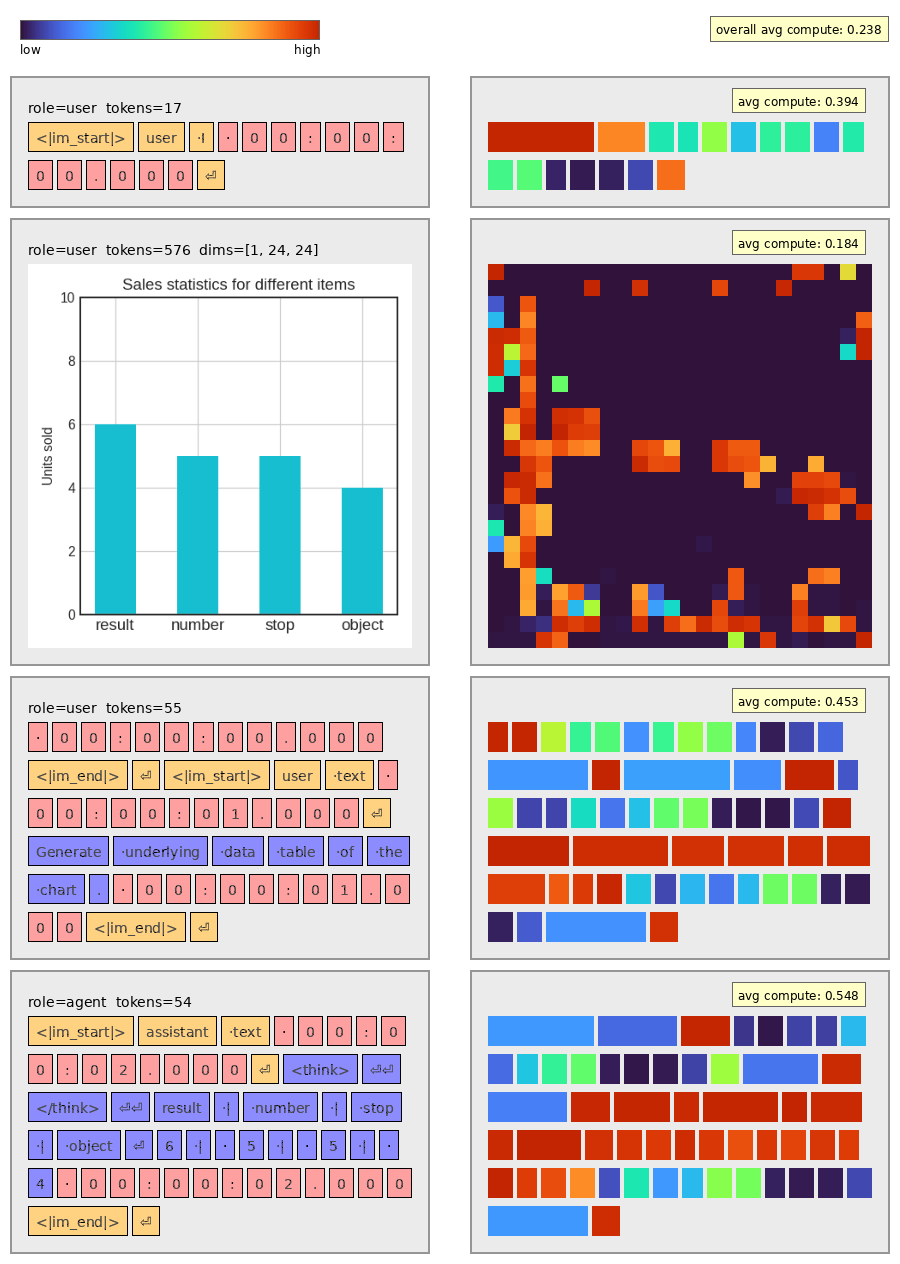}
  \caption{Copy experts produce polarized routing. Tokens route fully to real experts (bright) or fully to copy experts (dark), with few intermediate values. Mixed routing causes residual dilution, incentivizing all-or-nothing decisions. Compare to zero expert compute maps (Figure~\ref{fig:compute_map_0}), which show smoother gradation.}
  \label{fig:compute_map_copy_expert}
\end{figure}
}

\newcommand{\FigureComputeMapDocVQA}{
\begin{figure}[H]
  \centering
  \includegraphics[width=0.49\linewidth]{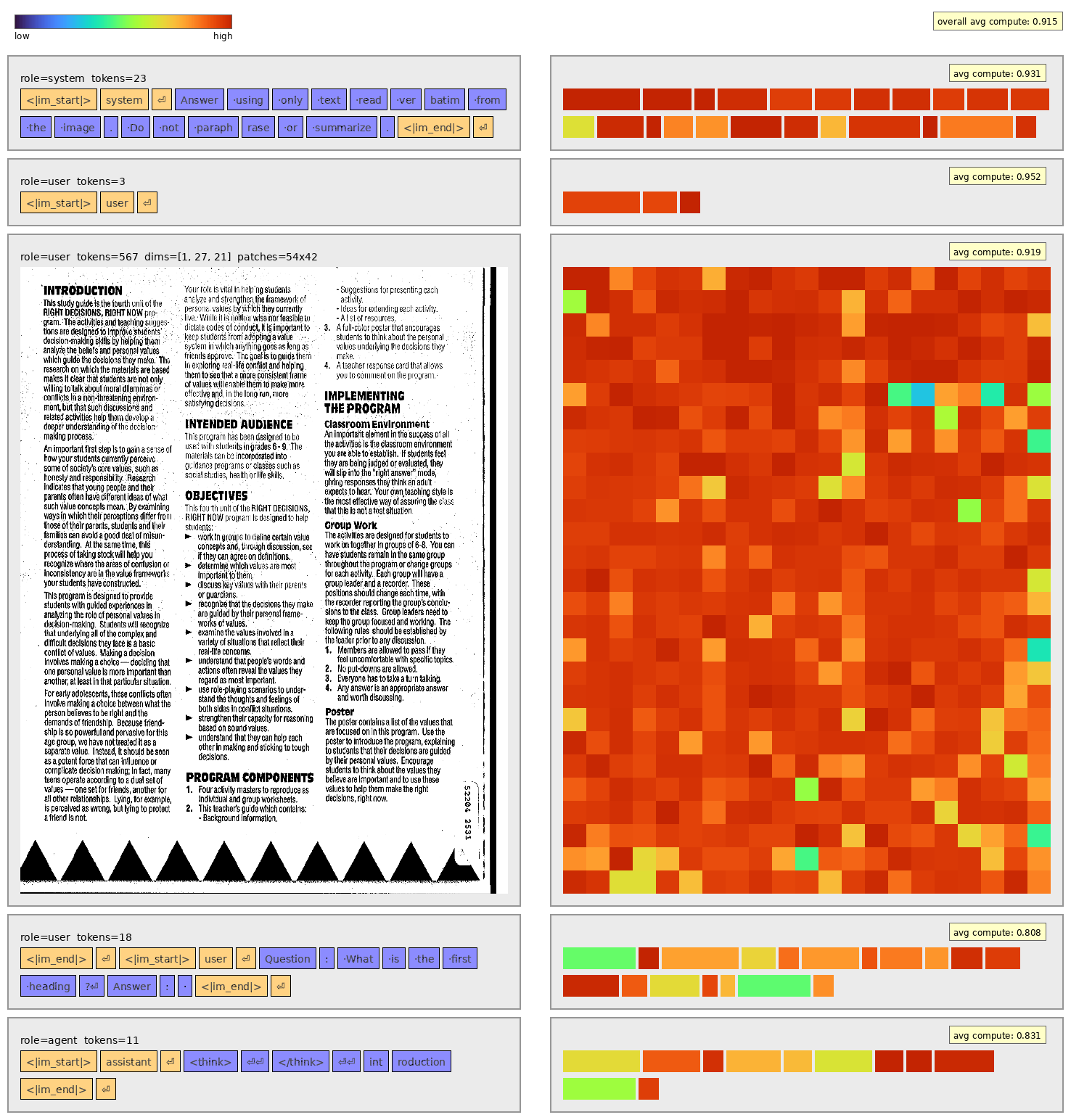}
  \hfill
  \includegraphics[width=0.49\linewidth]{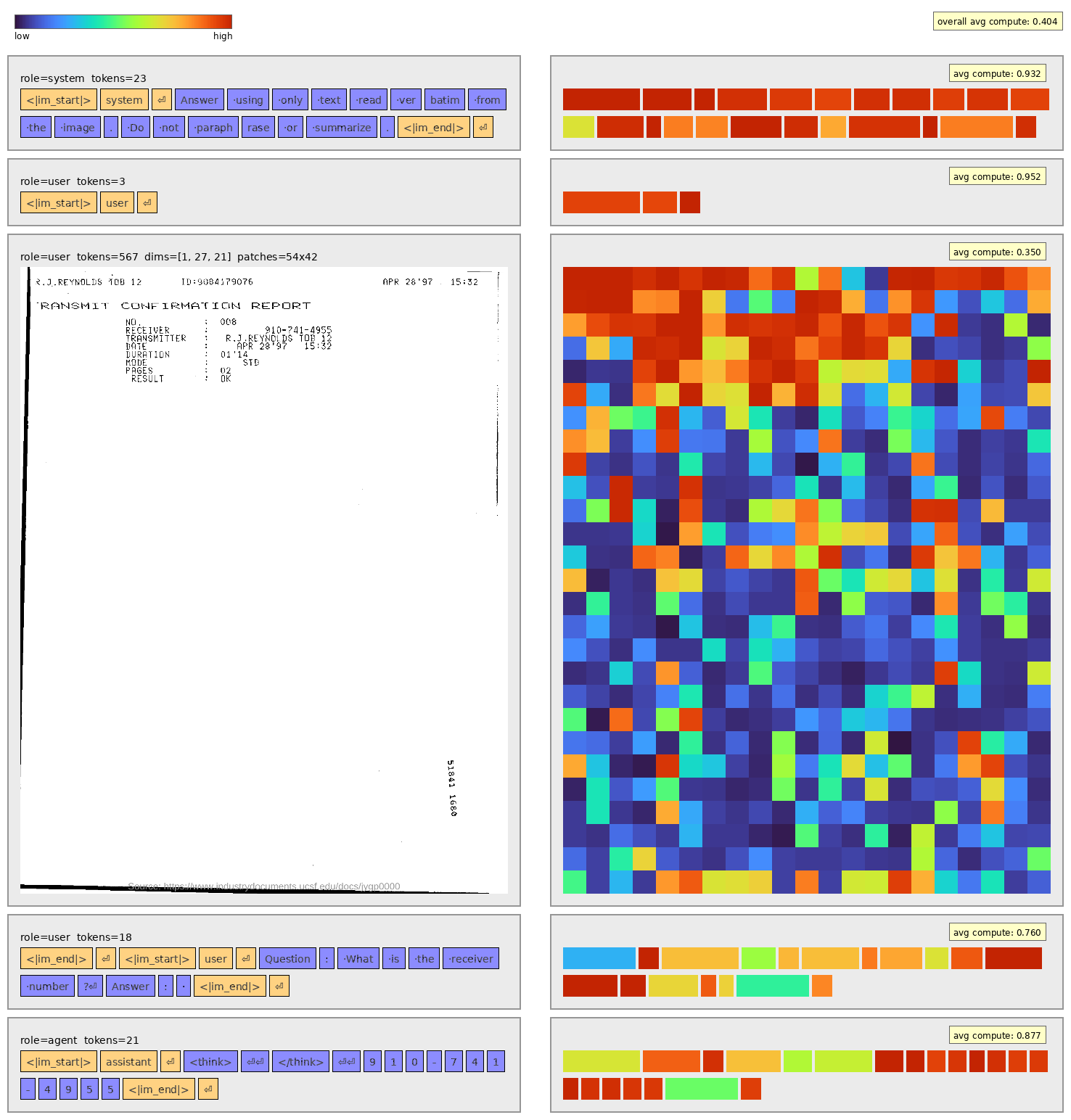}
  \caption{DocVQA samples elicit diverse compute maps due to task specificity and importance of fine-grained features. \textbf{Left:} A sample with lots of text which can be relevant to the question. The model allocates a lot of compute to understand each patch. \textbf{Right:} A sample with lots of white space which the model can save compute on. The sample on the left uses more than double average compute.}
  \label{fig:compute_map_docvqa}
\end{figure}
}

\newcommand{\FigureComputeMapLayers}{
\begin{figure}[H]
  \centering
  \includegraphics[width=0.8\linewidth]{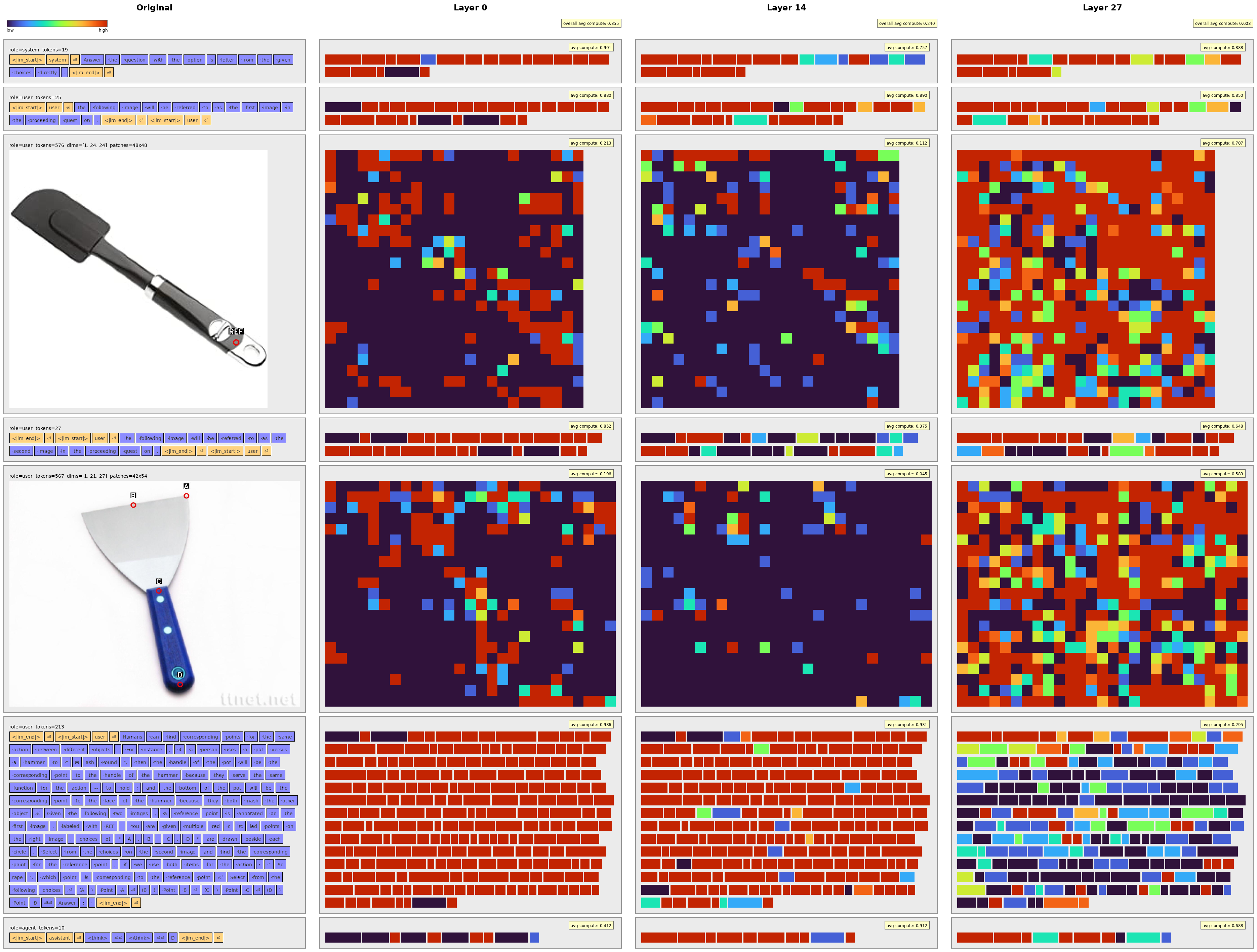}
  \\[0.5em]
  \includegraphics[width=0.8\linewidth]{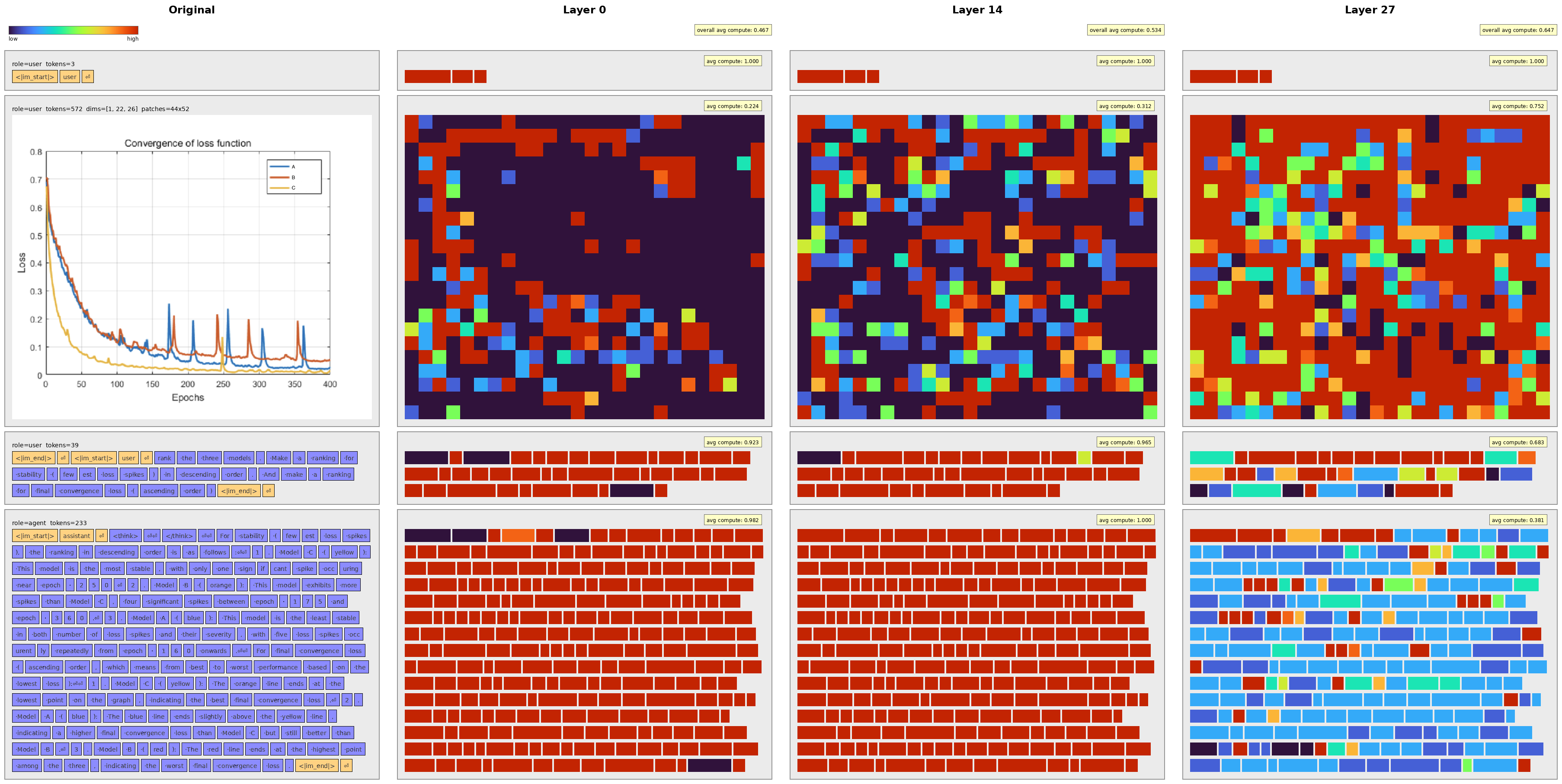}
  \\[0.5em]
  \includegraphics[width=0.8\linewidth]{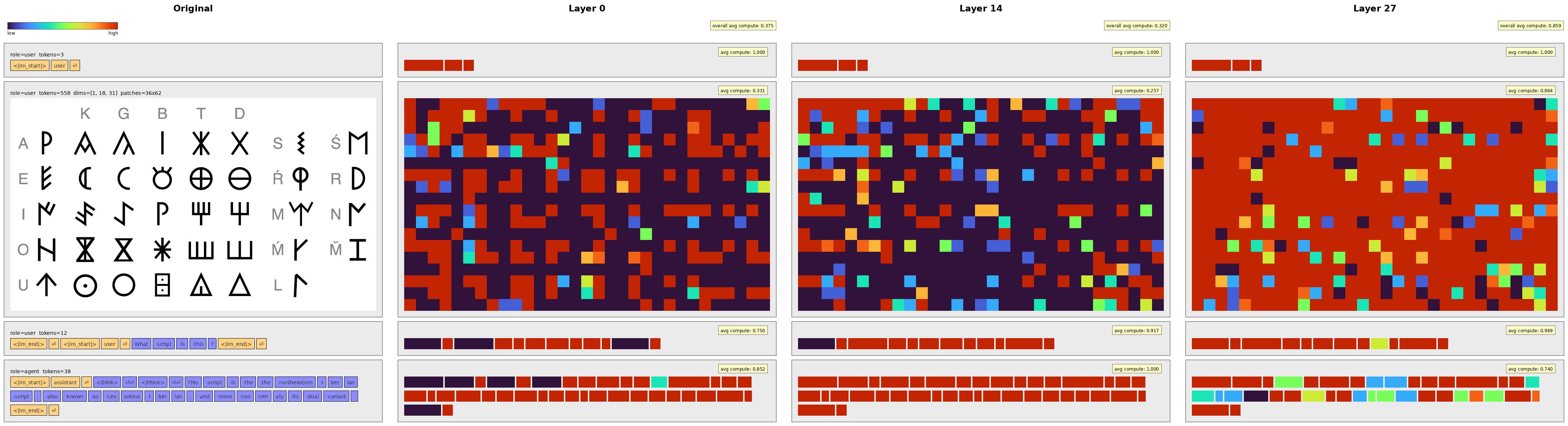}
  \caption{Compute allocation strategy varies significantly between layers. Earlier layers are much more focused on text and semantic aspects of the image whereas the last layer focuses more on the entirety of the image, allocating less compute to text.}
  \label{fig:compute_map_layers}
\end{figure}
}

\definecolor{plotblue}{RGB}{37,99,235}

\tikzset{
  gridline/.style={black!70, line width=0.5pt},
  labelsmall/.style={font=\scriptsize},
  labeltiny/.style={font=\tiny, gray},
  budgetlabel/.style={font=\tiny, gray!70},
  token/.style={draw, rounded corners=1pt, minimum width=0.5cm, minimum height=0.3cm, font=\tiny},
  tokenactive/.style={draw, rounded corners=1pt, minimum width=0.5cm, minimum height=0.3cm, font=\tiny, fill=plotblue!35},
  router/.style={draw, circle, minimum size=0.4cm, font=\tiny, fill=plotblue!30},
  expert/.style={draw, rectangle, minimum width=0.45cm, minimum height=0.3cm, font=\tiny},
  expertactive/.style={draw, rectangle, minimum width=0.45cm, minimum height=0.3cm, font=\tiny, fill=plotblue!45},
  nullexpert/.style={draw, dashed, rectangle, minimum width=0.45cm, minimum height=0.3cm, font=\tiny}
}

\newcommand{\WeightDataSparsityTikz}{
  \begin{tikzpicture}[x=0.65cm,y=0.65cm]

    \begin{scope}[xshift=0cm]
      \fill[black!8] (0,0) rectangle (3,3);

      \fill[black!25] (1,2) rectangle (2,3);
      \fill[black!25] (2,2) rectangle (3,3);
      \fill[black!25] (0,1) rectangle (1,2);
      \fill[black!25] (2,1) rectangle (3,2);
      \fill[black!25] (0,0) rectangle (1,1);
      \fill[black!25] (2,0) rectangle (3,1);

      \fill[plotblue!50] (0,3) rectangle (1,4);
      \fill[plotblue!50] (1,3) rectangle (2,4);

      \draw[gridline] (0,0) rectangle (3,4);
      \foreach \x in {1,2} \draw[gridline] (\x,0) -- (\x,4);
      \foreach \y in {1,2,3} \draw[gridline] (0,\y) -- (3,\y);

      \node[budgetlabel] at (4.0,3.5) {$\leq \rho_w E$};
      \draw[decorate, decoration={brace, amplitude=2pt, mirror}, gray!70]
        (3.15,3) -- (3.15,4);

      \node[font=\scriptsize] at (1.5,5.0) {Weight sparsity};
      \node[font=\tiny, gray] at (1.5,4.4) {(per-token budget)};

      \node[font=\tiny, rotate=90] at (-0.6,2) {tokens};
      \node[font=\tiny] at (1.5,-0.6) {experts};
    \end{scope}

    \node[font=\Large] at (3.4cm,1.3cm) {$\times$};

    \begin{scope}[xshift=4.9cm]
      \fill[black!8] (1,0) rectangle (3,4);

      \fill[black!25] (1,2) rectangle (2,3);
      \fill[black!25] (1,1) rectangle (2,2);
      \fill[black!25] (2,1) rectangle (3,2);
      \fill[black!25] (2,0) rectangle (3,1);

      \fill[plotblue!50] (0,1) rectangle (1,2);
      \fill[plotblue!50] (0,3) rectangle (1,4);

      \draw[gridline] (0,0) rectangle (3,4);
      \foreach \x in {1,2} \draw[gridline] (\x,0) -- (\x,4);
      \foreach \y in {1,2,3} \draw[gridline] (0,\y) -- (3,\y);

      \draw[decorate, decoration={brace, amplitude=2pt, mirror}, gray!70]
        (0,-0.2) -- (1,-0.2);
      \node[budgetlabel] at (0.5,-0.6) {$\leq \rho_d T$};

      \node[font=\scriptsize] at (1.5,5.0) {Data sparsity};
      \node[font=\tiny, gray] at (1.5,4.4) {(per-expert budget)};
    \end{scope}

    \node[font=\Large] at (8.2cm,1.3cm) {$=$};

    \begin{scope}[xshift=9.6cm]
      \fill[plotblue!50] (0,3) rectangle (1,4);
      \fill[plotblue!50] (1,3) rectangle (2,4);
      \fill[plotblue!50] (2,3) rectangle (3,4);
      \fill[plotblue!50] (0,2) rectangle (1,3);
      \fill[plotblue!50] (2,1) rectangle (3,2);

      \draw[gridline] (0,0) rectangle (3,4);
      \foreach \x in {1,2} \draw[gridline] (\x,0) -- (\x,4);
      \foreach \y in {1,2,3} \draw[gridline] (0,\y) -- (3,\y);

      \draw[decorate, decoration={brace, amplitude=3pt, mirror}, gray!70]
        (3.2,0) -- (3.2,4);
      \draw[decorate, decoration={brace, amplitude=3pt, mirror}, gray!70]
        (0,-0.2) -- (3,-0.2);
      \node[budgetlabel] at (3.6,2) {$T$};
      \node[budgetlabel] at (1.5,-0.6) {$E$};
      \node[budgetlabel] at (4.2,-0.6) {$\leq \rho_w \rho_d TE$};

      \node[font=\scriptsize] at (1.5,5.0) {Composed};
      \node[font=\tiny, gray] at (1.5,4.4) {(total budget)};
    \end{scope}

    \path (-1.6cm,0);
    \path (13.7cm,0);

  \end{tikzpicture}
}

\newcommand{\ImplementationsTikz}{
  \begin{tikzpicture}[x=0.85cm,y=0.85cm]

    \begin{scope}[xshift=0cm]
      \draw[dash pattern=on 2pt off 2pt, dash phase=0pt, gray!50] (0.15,-0.55) -- (0.15,0.65);
      \node[labeltiny] at (0.15,0.85) {now};
      \node[tokenactive] (t1) at (0.15,0) {$t_1$};
      \node[token, gray!50] (t2) at (0.82,0) {$t_2$};
      \node[token, gray!50] (t3) at (1.49,0) {$t_3$};
      \node[token, gray!50] (t4) at (2.16,0) {$t_4$};
      \node[router] (r1) at (0.15,-1.1) {R};
      \node[expertactive] (e1) at (0.75,-2.0) {$E_1$};
      \node[expertactive] (e2) at (1.49,-2.0) {$E_2$};
      \node[expert] (e3) at (2.23,-2.0) {$E_3$};
      \draw[->, plotblue!80] (t1) -- (r1);
      \draw[->, plotblue!80] (r1) -- (e1);
      \draw[->, plotblue!80] (r1) -- (e2);
      \node[labelsmall] at (1.15,1.5) {Token-choice};
      \node[labeltiny] at (1.15,1.15) {(weight sparsity, causal)};
      \node[labeltiny] at (1.49,-2.45) {fixed \textit{k}=2};
    \end{scope}

    \begin{scope}[xshift=4.9cm]
      \draw[dash pattern=on 2pt off 2pt, dash phase=0pt, gray!50] (1.49,-0.55) -- (1.49,0.65);
      \node[labeltiny] at (1.49,0.85) {now};
      \node[tokenactive] (u1) at (0.15,0) {$t_1$};
      \node[token] (u2) at (0.82,0) {$t_2$};
      \node[tokenactive] (u3) at (1.49,0) {$t_3$};
      \node[token] (u4) at (2.16,0) {$t_4$};
      \node[router] (r1b) at (0.82,-1.1) {R};
      \node[expertactive] (e1b) at (0.75,-2.0) {$E_1$};
      \node[expert] (e2b) at (1.49,-2.0) {$E_2$};
      \node[expert] (e3b) at (2.23,-2.0) {$E_3$};
      \draw[->, plotblue!80] (e1b) -- (r1b);
      \draw[->, plotblue!80] (r1b) -- (u1);
      \draw[->, plotblue!80] (r1b) -- (u3);
      \node[labelsmall] at (1.15,1.5) {Expert-choice};
      \node[labeltiny] at (1.15,1.15) {(data sparsity, non-causal)};
      \node[labeltiny] at (1.49,-2.45) {fixed \textit{c}=2};
    \end{scope}

    \begin{scope}[xshift=9.6cm]
      \draw[dash pattern=on 2pt off 2pt, dash phase=0pt, gray!50] (1.49,-0.55) -- (1.49,0.65);
      \node[labeltiny] at (1.49,0.85) {now};
      \node[token] (v1) at (0.15,0) {$t_1$};
      \node[token] (v2) at (0.82,0) {$t_2$};
      \node[tokenactive] (v3) at (1.49,0) {$t_3$};
      \node[token, gray!50] (v4) at (2.16,0) {$t_4$};
      \node[router] (r2) at (1.49,-1.1) {R};
      \node[expert] (re1) at (0.01,-2.0) {$E_1$};
      \node[expert] (re2) at (0.75,-2.0) {$E_2$};
      \node[expertactive] (re3) at (1.49,-2.0) {$E_3$};
      \node[nullexpert] (ne1) at (2.23,-2.0) {$\varnothing$};
      \node[nullexpert] (ne2) at (2.97,-2.0) {$\varnothing$};
      \draw[->, plotblue!80] (v3) -- (r2);
      \draw[->, plotblue!80] (r2) -- (re3);
      \draw[->, dashed, gray!60] (r2) -- (ne1);
      \node[labelsmall] at (1.49,1.5) {Null experts (ours)};
      \node[labeltiny] at (1.49,1.15) {(composed, causal)};
      \node[labeltiny] at (1.49,-2.45) {\textit{k}=2, variable real};
    \end{scope}

    \path (-1.6cm,0);
    \path (13.7cm,0);

  \end{tikzpicture}
}

\newcommand{\FigMethodOverview}{
\begin{figure}[t]
  \centering
  \resizebox{\linewidth}{!}{\WeightDataSparsityTikz}
  \vspace{0.6em}
  \resizebox{\linewidth}{!}{\ImplementationsTikz}
  \caption{\textbf{Top:} Weight and data sparsity as dual budget constraints. Weight sparsity bounds experts per token (row budget $\leq \rho_w E$); data sparsity bounds tokens per expert (column budget $\leq \rho_d T$); composing both yields a budget over the full $T \times E$ matrix. \textbf{Bottom:} Implementations. Token-choice achieves weight sparsity causally; expert-choice achieves data sparsity but requires seeing future tokens; null experts compose both while preserving causality.}
  \label{fig:method_overview}
\end{figure}
}